\documentclass[10pt,twocolumn,letterpaper]{article}
\usepackage{cvpr}
\usepackage{times}
\usepackage{epsfig}
\usepackage{graphicx}
\usepackage{amsmath}
\usepackage{amssymb}
\usepackage{multicol}
\usepackage{multirow}
\usepackage{booktabs}
\usepackage{mathtools}
\usepackage[LGR,T1]{fontenc}
\DeclareSymbolFont{upgreek}{LGR}{cmr}{m}{n}
\SetSymbolFont{upgreek}{bold}{LGR}{cmr}{bx}{n}
\DeclareMathSymbol{\uplambda}{\mathord}{upgreek}{`l}
\def\ourmethod{MUXNet}

\def\Minimize{\mathop{\rm minimize}\limits}

\def\st{\mathop{\rm subject\ to}}
\DeclarePairedDelimiterX{\norm}[1]{\lVert}{\rVert}{#1}

\usepackage{bm}
\usepackage{pdfpages}
\usepackage[font=footnotesize,labelfont=bf]{caption}
\usepackage{subcaption}
\usepackage[pagebackref=true,breaklinks=true,letterpaper=true,colorlinks,bookmarks=false]{hyperref}
\usepackage[page]{appendix}

\newcommand{\nobyno}[1]{\mbox{$#1\!\!\times\!\!#1$}}

\cvprfinalcopy
\ifcvprfinal\fi

\begin{document}
\title{MUXConv: Information Multiplexing in Convolutional Neural Networks}
\author{Zhichao Lu\quad\quad Kalyanmoy Deb\quad\quad Vishnu Naresh Boddeti\\
Michigan State University\\
{\tt\small \{luzhicha, kdeb, vishnu\}@msu.edu}
}

\maketitle

\begin{abstract}
    Convolutional neural networks have witnessed remarkable improvements in computational efficiency in recent years. A key driving force has been the idea of trading-off model expressivity and efficiency through a combination of \nobyno{1}\ and depth-wise separable convolutions in lieu of a standard convolutional layer. The price of the efficiency, however, is the sub-optimal flow of information across space and channels in the network. To overcome this limitation, we present MUXConv, a layer that is designed to increase the flow of information by progressively multiplexing channel and spatial information in the network, while mitigating computational complexity. Furthermore, to demonstrate the effectiveness of MUXConv, we integrate it within an efficient multi-objective evolutionary algorithm to search for the optimal model hyper-parameters while simultaneously optimizing accuracy, compactness, and computational efficiency. On ImageNet, the resulting models, dubbed \ourmethod{}s, match the performance (75.3\% top-1 accuracy) and multiply-add operations (218M) of MobileNetV3 while being 1.6$\times$ more compact, and outperform other mobile models in all the three criteria. \ourmethod{} also performs well under transfer learning and when adapted to object detection. On the ChestX-Ray 14 benchmark, its accuracy is comparable to the state-of-the-art while being $3.3\times$ more compact and $14\times$ more efficient. Similarly, detection on PASCAL VOC 2007 is 1.2\% more accurate, 28\% faster and 6\% more compact compared to MobileNetV2. The code is available from \url{https://github.com/human-analysis/MUXConv}.
\end{abstract}


\section{Introduction}

\begin{figure}[t]
	\centering
	\begin{subfigure}[t]{.47\textwidth}
		\centering
		\includegraphics[width=0.98\textwidth]{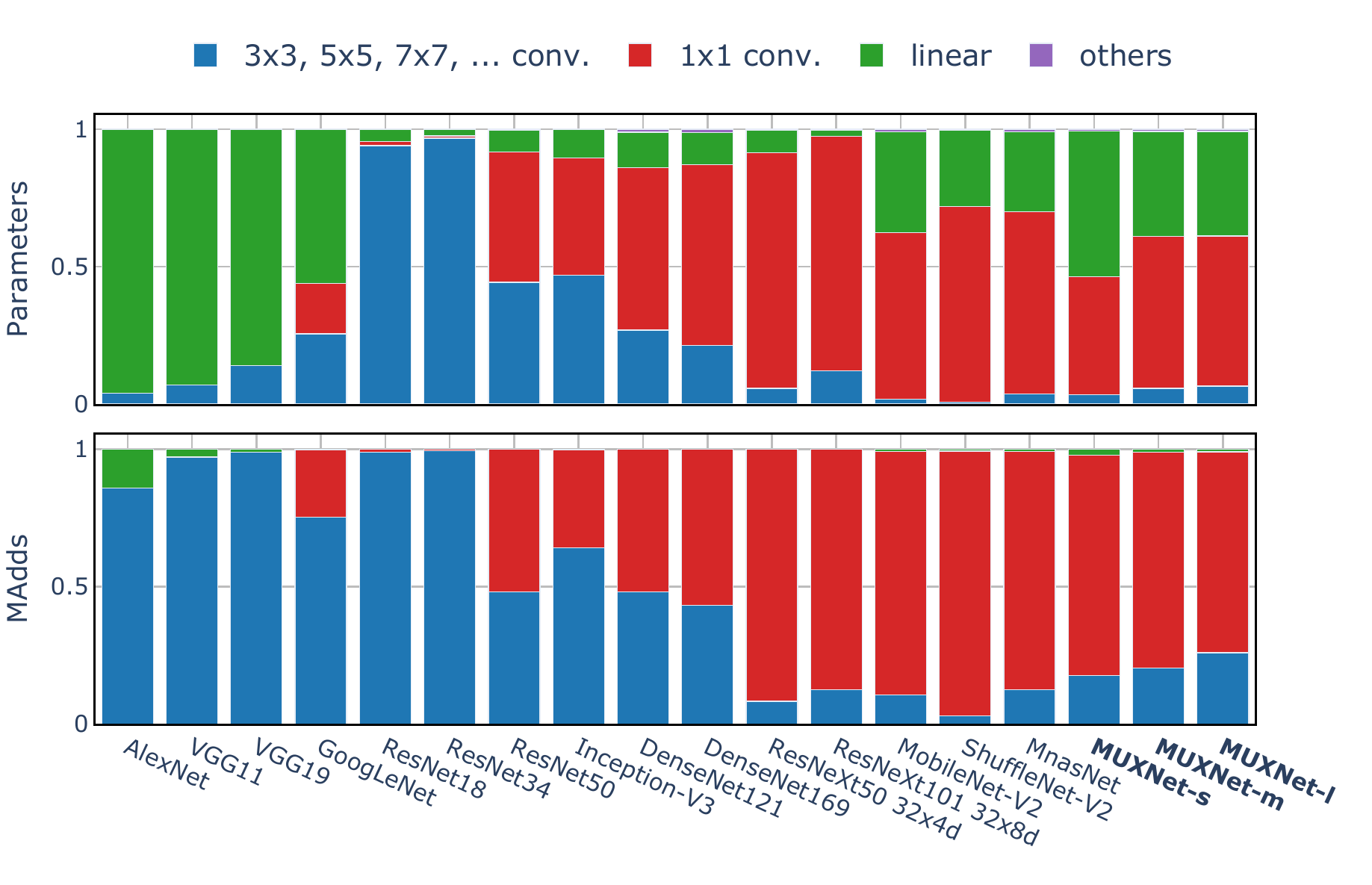}
	\end{subfigure}\\
	\caption{Relative contribution of different layers in CNN designs in terms of parameters (top) and MAdds (bottom). Initial models largely relied on standard convolutional layers. More recent networks, on the other hand, largely rely on \nobyno{1}\ convolutions and linear layers. In contrast, MUXNets reverse this trend to an extent.\label{fig:complexity_distribution}\vspace{-0.2cm}}
\end{figure}

In the span of the last decade, convolutional neural networks (CNNs) have undergone a dramatic transformation in terms of predictive performance, compactness and computational efficiency. The development largely happened in two phases. Starting from AlexNet~\cite{alexnet}, the focus of the first wave of models was on improving the predictive accuracy of CNNs including VGG~\cite{vgg}, GoogleNet~\cite{googlenet}, ResNet~\cite{resnet}, ResNeXt~\cite{resnext}, DenseNet~\cite{densenet} etc. These models progressively increased the contribution of \nobyno{3}\ convolutions, both in model size as well as multiply-add operations (MAdds). The focus of the second wave of models was on improving their computational efficiency while trading-off accuracy to a small extent. Models in this category include ShuffleNet~\cite{shufflenetv2}, MobileNetV2~\cite{mobilenetv2}, MnasNet~\cite{mnasnet} and MobileNetV3~\cite{mobilenetv3}. Such solutions sought to improve computational efficiency by progressively replacing the parameter and compute intensive standard convolutions by a combination of \nobyno{1}\ convolutions and depth-wise separable \nobyno{3}\ convolutions. Figure \ref{fig:complexity_distribution} depicts the trend in the relative contributions of different layers in terms of parameters and MAdds.
\begin{figure}[t]
	\centering
	\begin{subfigure}[t]{.47\textwidth}
		\centering
		\includegraphics[width=0.98\textwidth, trim=0 1cm 0 0, clip=true]{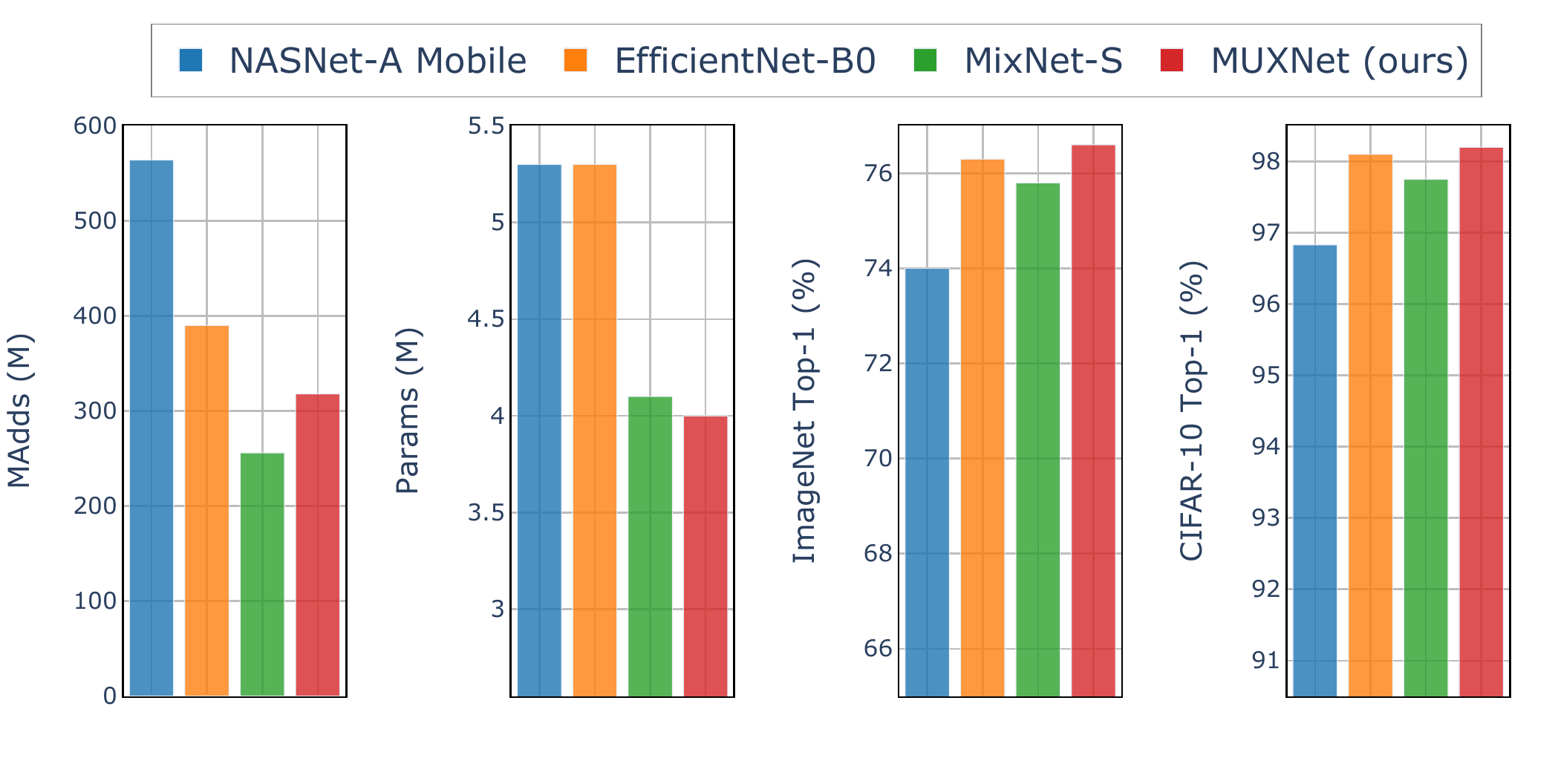}
	\end{subfigure}\\
	\begin{subfigure}[t]{.47\textwidth}
		\centering
		\includegraphics[width=0.98\textwidth]{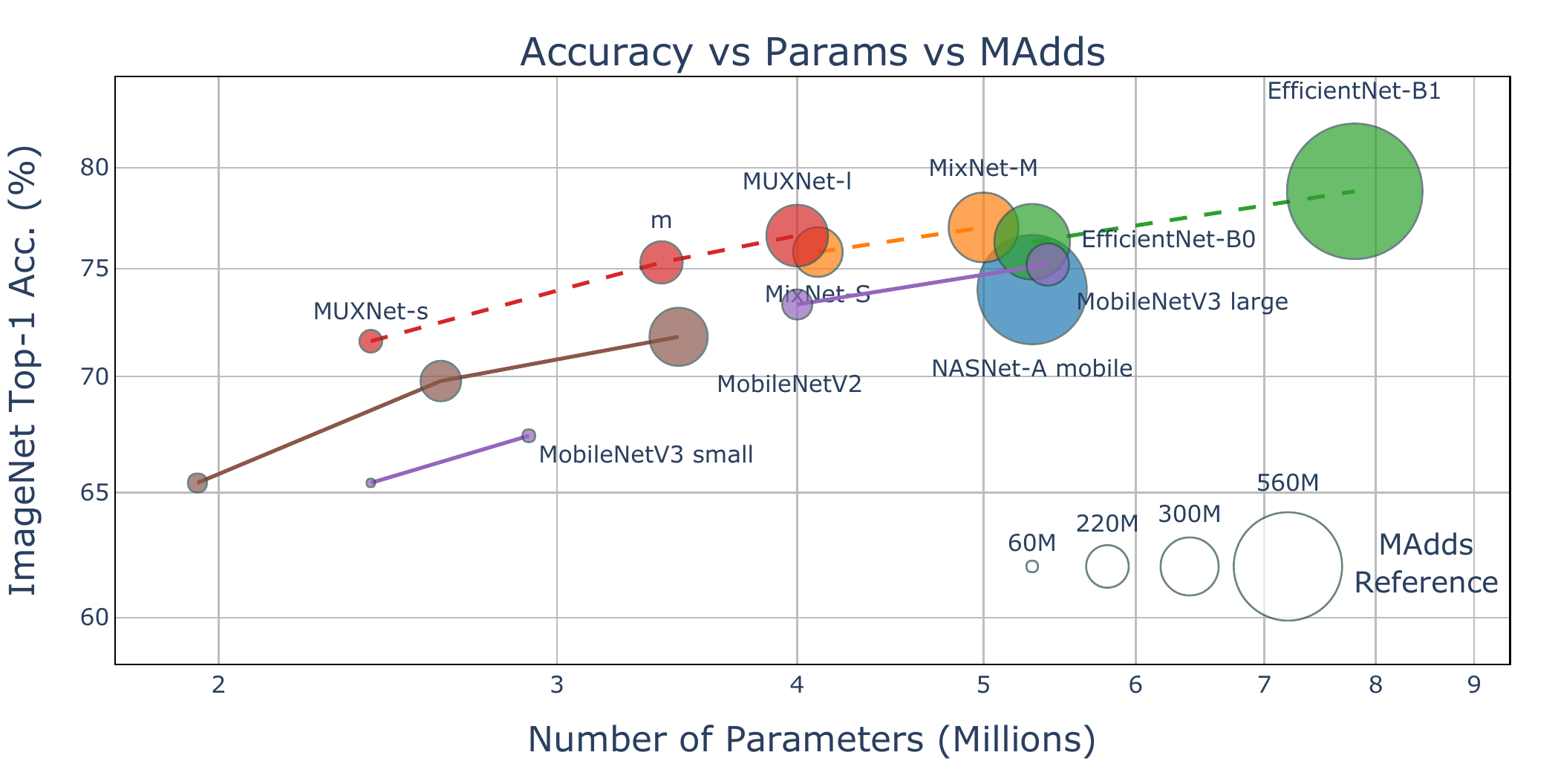}
	\end{subfigure}\\
	\caption{\textbf{Accuracy vs. Compactness vs. Efficiency:} Existing networks outperform each other in at most two criteria. MUXNet models are, however, dominant in all three objectives under mobile settings.
	\label{fig:preview_results}\vspace{-0.2cm}}
\end{figure}

Depth-wise separable convolutions~\cite{sifre2014rigid,chollet2017xception} offer significant computational benefits, both from the perspective of number of parameters as well as computational complexity. A salient feature of these layers is the lack of interactions between information in the channels. This limitation is overcome through \nobyno{1} convolution, a layer which allows for interactions and information flow across the channels. The combination of depth-wise separable and \nobyno{1}\ convolution fully decouples the task of spatial and channel information flow, respectively, into two independent and efficient layers. On the other hand, a standard convolutional layer couples the spatial and channel information flow into a single, yet, computationally inefficient layer. Therefore, the former replaced the latter as the workhorse of CNN designs.

In this paper, we seek an alternative approach to trade-off the expressivity and efficiency of convolutional layers. We introduce MUXConv, a layer that leverages the efficiency of depth-wise or group-wise convolutional layers along with a mechanism to enhance the flow of information in the network. MUXConv achieves this through two components, spatial multiplexing and channel multiplexing. Spatial multiplexing extracts feature information at multiple scales via spatial shuffling, processes such information through depth-wise or group-wise convolutions and then unshuffles them back together. Channel multiplexing is inspired by ShuffleNet~\cite{shufflenetv2} and is designed to address the limitation of depth-wise/group convolutions, namely the lack of information flow across channels/groups of channels, by shuffling the channels. The shuffling procedure and the operations we perform on the shuffled channels are motivated by computational efficiency and differ significantly from ShuffleNet. Collectively, these two components increase the flow of information, both spatially and across channels, while mitigating the computational burden of the layer.

To further realize the full potential of MUXConv in trading-off accuracy and computational efficiency, we propose a population based evolutionary algorithm to efficiently search for the hyperparameters of each MUXConv layer in the network. The search simultaneously optimizes three objectives, namely, prediction accuracy, model compactness and model efficiency in terms of MAdds. To improve the efficiency of the search process we decompose the multi-objective optimization problem into a collection of single-objective optimization sub-problems, that are in turn optimized simultaneously and cooperatively. We refer to the resulting family of CNNs as \ourmethod{}s.

\vspace{3pt}
\noindent\textbf{Contributions:} We first develop a new layer, called MUXConv, that multiplexes information flow spatially and across channels while improving the computational efficiency of equivalent combination of depth-wise separable and \nobyno{1}\ convolutions. Then, we develop the first multi-objective neural architecture search (NAS) algorithm to simultaneously optimize compactness, efficiency, and accuracy of \ourmethod{}s designed with MUXConv as the basic building block. We present thorough experimental evaluation demonstrating the efficacy and value of each component of \ourmethod{} across multiple tasks including image classification (ImageNet), object detection (PASCAL VOC 2007) and transfer learning (CIFAR-10, CIFAR-100, ChestX-Ray14). Our results indicate that, unlike the conventional wisdom in all existing solutions, it is feasible to design CNNs that do not sacrifice compactness for efficiency or vice versa in the quest for better predictive performance.


\section{Related-work}\label{sec:related}
Many CNN architectures have been developed by optimizing different objectives, such as, model compactness, computational efficiency, or predictive performance. Below, we categorize the solutions into a few major themes.

\vspace{3pt}
\noindent\textbf{Multi-Scale and Shuffling:} The notion of multi-scale processing in CNNs has been utilized in different forms and in a variety of contexts. These include explicit processing of multi-resolution feature maps for object detection~\cite{cai2016unified,lin2017feature} and image classification~\cite{huang2018multi} and computational blocks with built-in multi-scale processing~\cite{biglittle,gao2019res2net}. The focus of these methods is predictive performance and hence towards large scale models. In contrast, multi-scale processing in MUXConv is motivated by enhancing information flow in small scale models deployed in resource constrained environments. Notably, MUXConv scales the feature maps through a pixel shuffling operation that is similar to subpixel convolution in~\cite{shi2016real}. The channel shuffling component of MUXConv is motivated by~\cite{shufflenet,shufflenetv2}.

\vspace{3pt}
\noindent\textbf{Mobile Architectures:} A number of CNN architectures have been developed for mobile settings. These include SqueezeNet~\cite{squeezenet}, MobileNet~\cite{mobilenet}, MobileNetV2~\cite{mobilenetv2}, MobileNetV3~\cite{mobilenetv3}, ShuffleNet~\cite{shufflenet}, ShuffleNetV2~\cite{shufflenetv2} and CondenseNet~\cite{condensenet}. The focus of this body of work has largely been to optimize two objectives, either accuracy and compactness or accuracy and efficiency, thereby resulting in models that are either efficient or compact but not both. In contrast, MUXNets are designed to simultaneously optimize all three objectives, compactness, efficiency and accuracy, and therefore leads to models that are both compact and efficient at the same time.

\vspace{3pt}
\noindent\textbf{Neural Architecture Search:} Automated approaches to search for good neural architectures have proven to be very effective in finding computational blocks that not only exhibit high predictive performance but also generalize and transfer to other tasks. Majority of the approaches including, NasNet~\cite{nasnet}, PNAS~\cite{PNAS}, DARTS~\cite{darts}, AmoebaNet~\cite{amoebanet} and MixNet~\cite{mixnet}, are optimized against a single objective, namely predictive performance. A couple of recent approaches, LEMONADE~\cite{LEMONADE}, NSGANet~\cite{NSGANet}, simultaneously optimize the networks against multiple objectives, including parameters, MAdds, latency, and accuracy. However, only results on small-scale datasets like CIFAR-10 are demonstrated in both approaches. Concurrently, a number of CNN architectures, such as ProxylessNAS~\cite{proxylessnas}, MnasNet~\cite{mnasnet}, ChamNet~\cite{chamnet} and FBNet~\cite{chamnet}, have been designed to target specific computing platforms such as mobile, CPU, and GPU. In contrast to the aforementioned NAS approaches, we adopt a hybrid search strategy where the basic computational block, MUXConv, is hand-designed while the hyper-parameters of each MUXConv layer in the network are searched through a population based evolutionary algorithm directly on a large scale dataset.


\section{Multiplexed Convolutions \label{sec:muxconv}}
The multiplexed convolution layer, called MUXConv, is a combination of two components: (1) spatial multiplexing which enhances the expressivity and predictive performance of the network, and (2) channel multiplexing which aids in reducing the computational complexity of the model.

\subsection{Spatial Multiplexing}
\begin{figure}[t]
	\centering
	\begin{subfigure}{.47\textwidth}
		\centering
		\includegraphics[width=0.98\textwidth]{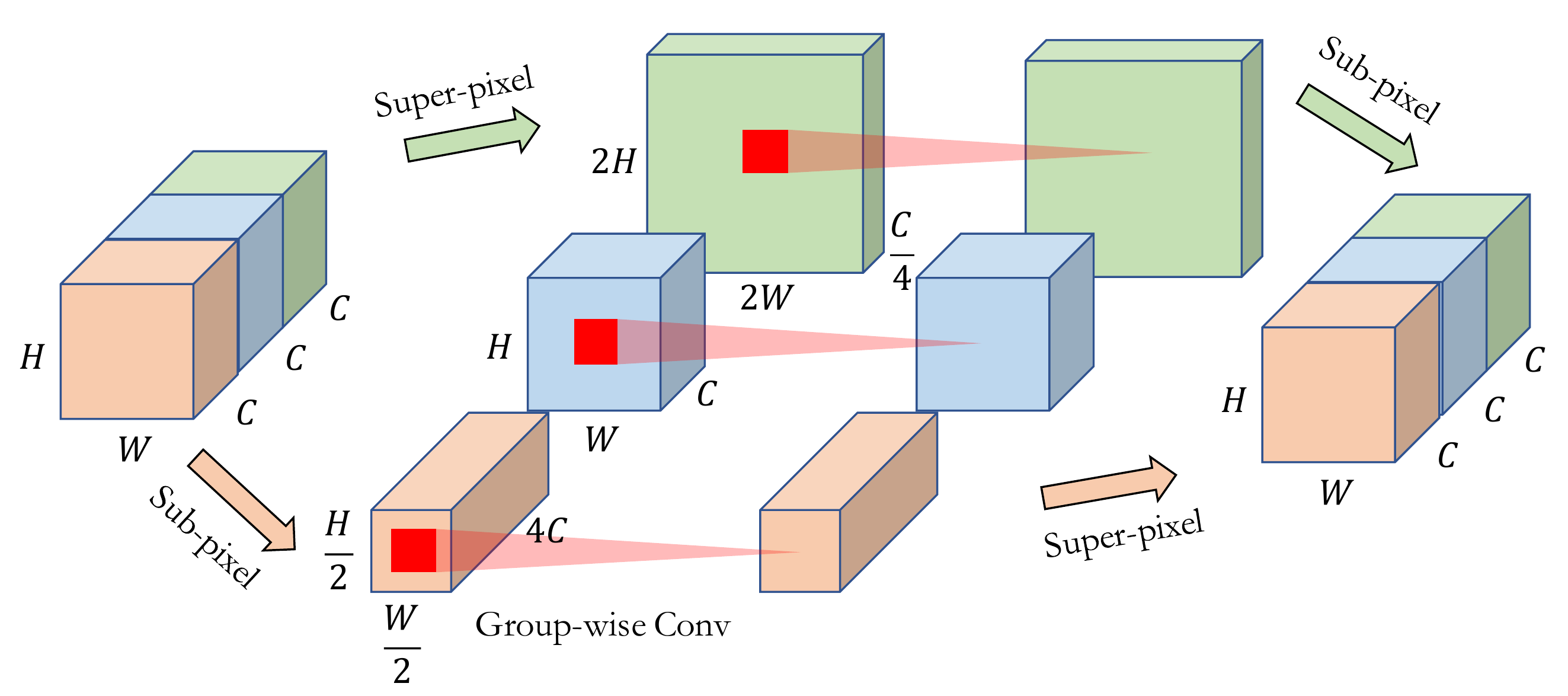}
		\vspace{-0.1cm}
		\caption{\label{fig:ms_block}}
	\end{subfigure}

	\begin{subfigure}[b]{.235\textwidth}
		\centering
		\includegraphics[width=0.98\textwidth]{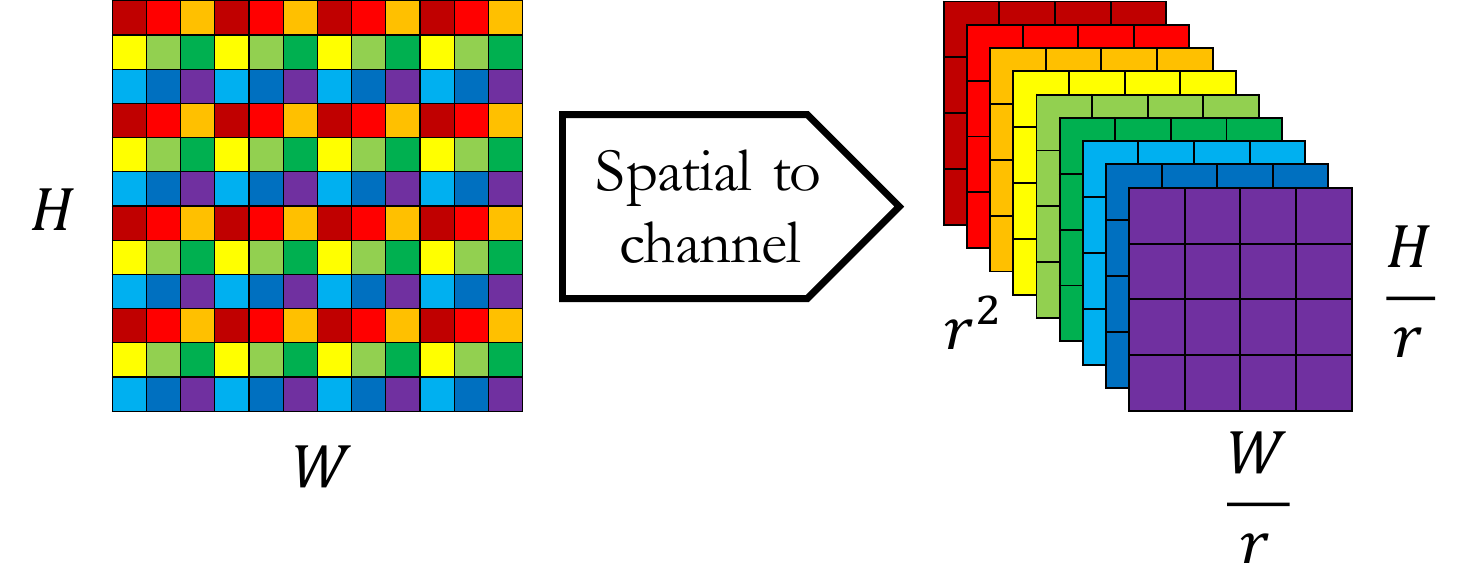}
		\vspace{-0.1cm}
		\caption{\label{fig:subpixel}}
	\end{subfigure}\hfill
	\begin{subfigure}[b]{.235\textwidth}
		\centering
		\includegraphics[width=0.98\textwidth, trim=0 5mm 0 0, clip=true]{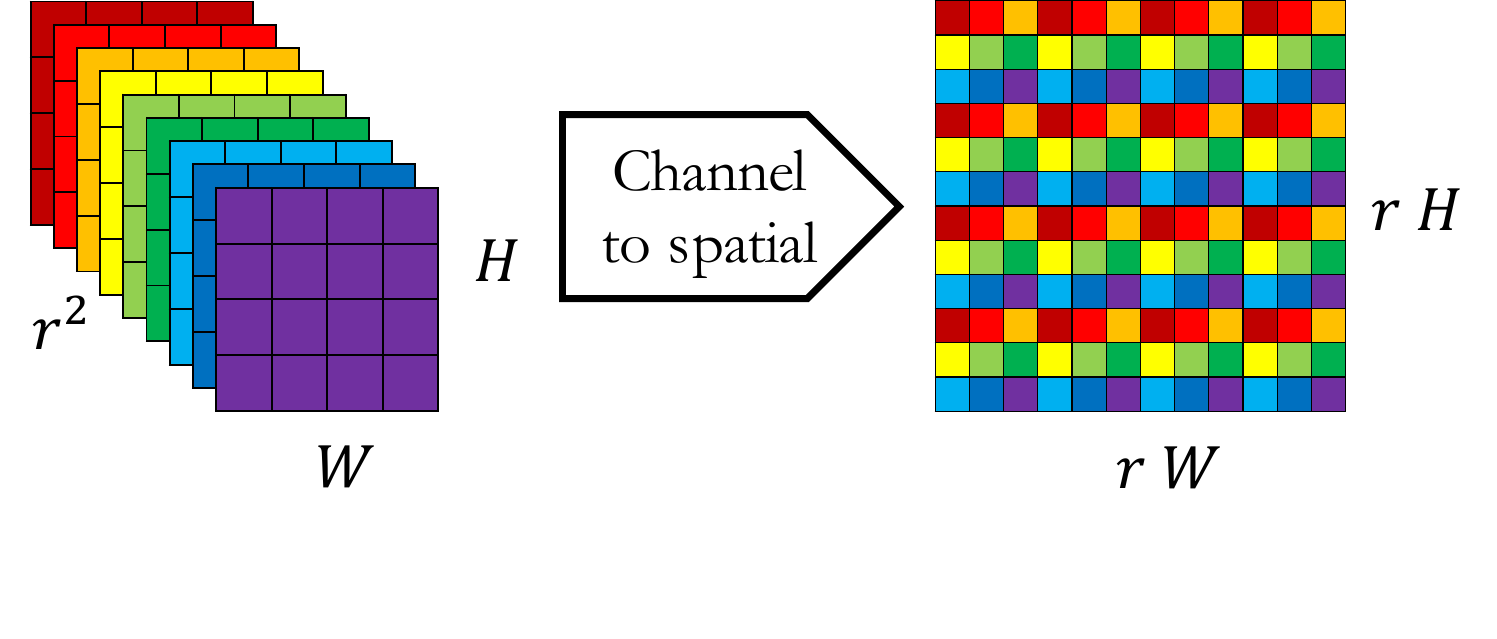}
		\vspace{-0.1cm}
		\caption{\label{fig:superpixel}}
	\end{subfigure}
	\caption{(a) Overview of spatial multiplexing operation. (b) Subpixel operation multiplexes spatial information into channels. (c) Superpixel operation multiplexes channels into spatial information.
	\label{fig:spatial_multiplexing}\vspace{-0.2cm}}
\end{figure}

The expressivity of a standard convolutional layer stems from the flow of information spatially and across the channels. Spatial multiplexing is designed to mimic this property while mitigating its computational complexity. The key idea is to map spatial information at multiple scales into channels and vice versa. Specifically, given a feature map $\bm{x} \in \mathbb{R}^{C\times H\times W}$, where $C$ is the number of channels, $H$ is the height and $W$ is the width of the feature map, the channels are grouped into three groups of $(C_1,C_2,C_3)$ channels such that $C=C_1+C_2+C_3$. The first and third group of channels are subjected to a subpixel and superpixel multiplexing operation, respectively. The multiplexed channels are then processed through a group-wise convolution operation defined over each of the three groups. The output feature maps from the group convolutions are mapped back to the same dimensions as the input feature maps by reversing the respective subpixel and superpixel operations. An illustration of this process is shown in Fig.~\ref{fig:ms_block}. Collectively, the subpixel and superpixel operations allow multi-scale spatial information to flow across channels. We note that the standard idea of multi-scale processing in existing approaches, multi-scale feature representations or kernels with larger receptive fields, is typically across different layers. In contrast, MUXConv seeks to exploit multi-scale information within a layer through pixel manipulation. As we show in Section~\ref{sec:ablation}, this operation significantly improves network accuracy especially as they get more compact.

We parameterize the subpixel multiplexing operation (see Fig. \ref{fig:subpixel}) by $r$ and define a window and stride of size \nobyno{r}. The features in the windows are mapped to $r^2$ channels, with each window corresponding to a unique feature location in the channels. On the whole, the subpixel operation maps the first group of channel features of size $C_1\times H\times W$ to features of size $r^2C_1\times\frac{H}{r}\times\frac{W}{r}$. Therefore, the subpixel operation enables down-scaled spatial information to be multiplexed with channel information and processed jointly by a standard convolution over the group. The combination of the two operations effectively increases the receptive field of the convolution by a factor of $r$.

We define the superpixel multiplexing operation (see Fig. \ref{fig:superpixel}) as an inverse of subpixel multiplexing. It is parameterized by $r^2$ which corresponds to the number of channels that will be multiplexed spatially into a single channel. The feature values at a particular location from the $r^2$ channels are mapped to a unique window in the output feature map. On the whole, the superpixel operation maps the third group of channels features of size $C_3\times H\times W$ to features of size $\frac{C_3}{r^2}\times rH\times rW$. Therefore, the superpixel operation enables channel information to be multiplexed with up-scaled spatial information and processed jointly by a standard convolution over the group. The combination of the two operations effectively decreases the receptive field of the convolution by a factor of $r$. Our superpixel operation bears similarity to the concept of \emph{tiled convolution}~\cite{ngiam2010tiled}, a particular realization of locally connected layers. This idea has also been particularly effective for image super-resolution~\cite{shi2016real} in the form of ``subpixel" convolution.

\begin{figure}[t]
	\centering
	\includegraphics[width=0.45\textwidth]{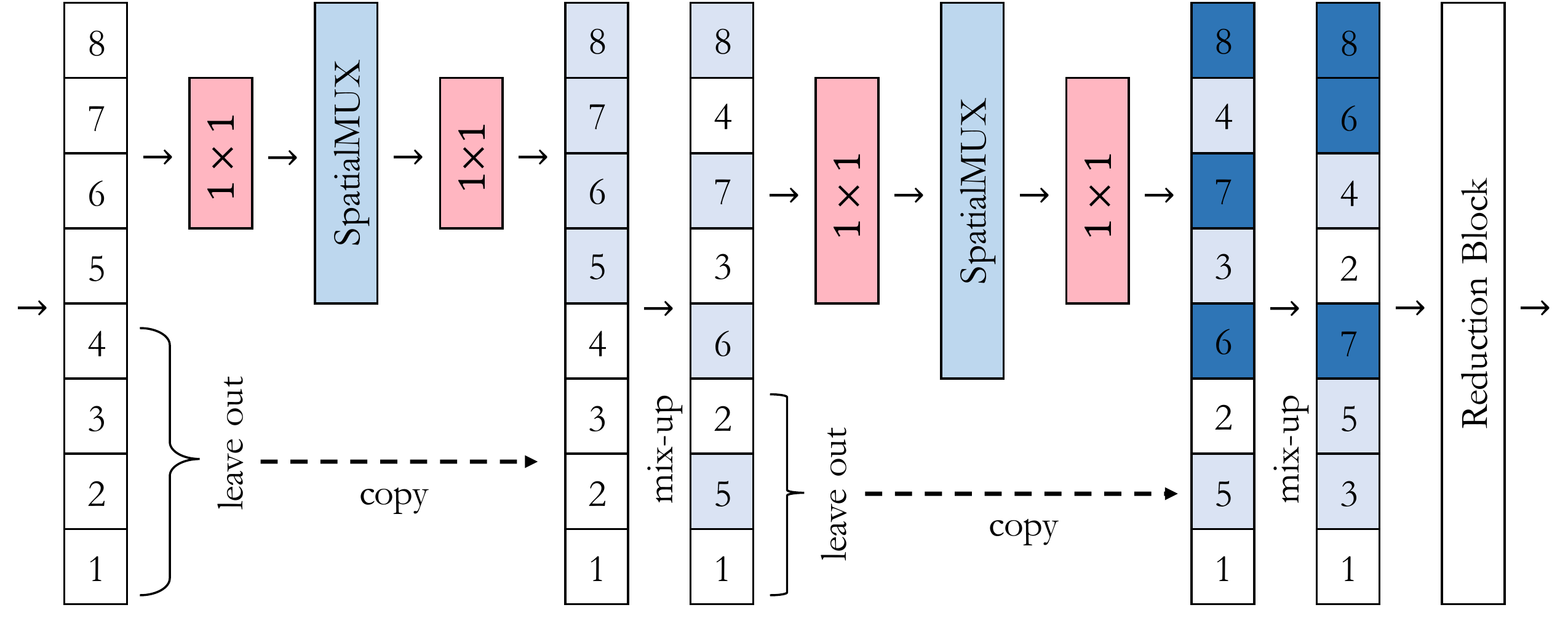}
	\caption{Illustration of two channel multiplexing layers. In each layer, half the channels are propagated as is while the other half are processed through the spatial multiplexing operation. The channels from the two groups are then interleaved as denoted by the indices. Color intensity denotes number of times that channel is processed.\label{fig:our_block}\vspace{-0.2cm}}
\end{figure}
\subsection{Channel Multiplexing}
While the spatial multiplexing operation described above is effective, it still suffers from some limitations. Firstly, the group convolutions in spatial multiplexing are more computationally expensive than depth-wise separable convolutions that they replace. Secondly, the decoupled nature of the group convolutions does not allow for flow of information across the groups. The channel multiplexing operation is designed to mitigate these drawbacks by reducing the computational burden of spatial multiplexing and further enhancing the flow of information across the feature map channels. This is achieved in two stages, selective processing and channel shuffling. A illustration of the whole operation is shown in Fig. \ref{fig:our_block}. Overall, the channel multiplexing operation is similar in spirit to ShuffleNet~\cite{shufflenet} and ShuffleNetV2~\cite{shufflenetv2} but with notable variations; (1) ShuffleNet uses shuffling to share channel information that are processed in different groups, while we use shuffling to blend the raw and processed channel information., (2) While ShuffleNetV2 always splits the input channels in half, we treat it as a hyperparameter that is searched for each layer, and (3) Shuffled channels are processed through an inverted residual bottleneck block in ShuffleNetV2 as opposed to spatial multiplexing in our case.

\vspace{5pt}
\noindent\textbf{Selective Processing:} We process only a part of the input channels by the spatial multiplexing block. Specifically, the $C$ channels in the input feature maps are split into two groups with $C_1$ and $C_2$ channels, such that $C=C_1+C_2$. The first group of channels are propagated as is while the second group are processed through spatial multiplexing. This scheme immediately increases the compactness and efficiency by a factor of $\left(\frac{C}{C_2}\right)^2$, which can compensate for the computational burden of grouped as opposed to depth-wise separable convolutions.

\vspace{5pt}
\noindent\textbf{Channel Shuffling:} After the selective processing operation, we shuffle the channels of the output feature map in a fixed pattern. Alternative channels selected from the unprocessed and processed channels are interleaved.

\section{Tri-Objective Hyperparameter Search \label{sec:moead}}
Designing a CNN typically involves many hyperparameters that critically impact the performance of the models. In order to realize the full potential of MUXNet we seek to search for the optimal hyperparameters in each layer of the network. Since the primary design motive of MUXConv is to increase model expressivity while mitigating computational complexity, we propose a multi-objective hyperparameter search algorithm to simultaneously optimize for accuracy, compactness and efficiency. This can be stated as,
\begin{equation}
    \begin{aligned}
    \Minimize_{} & \hspace{3mm} \textbf{F}(\bm{x}) = \big(f_1(\bm{x}), \cdots, f_m(\bm{x})\big)^T, \\
    \st & \hspace{3mm} \bm{x} \in \bm{\Omega},
    \end{aligned}
    \label{def:mop}
\end{equation}
\noindent where in our context $\bm{\Omega} = \Pi_{i=1}^{n}[a_i, b_i] \subseteq \mathbb{R}^n$ is the hyperparameter decision space, where $a_i$, $b_i$ are the lower and upper bounds, $\bm{x} = (x_1, \ldots, x_n)^T \in \bm{\Omega}$ is a candidate hyperparameter setting, \textbf{F} : $\bm{\Omega} \rightarrow \mathbb{R}^m$ constitutes $m$ competing objectives, i.e. predictive error, model size, model inefficiency, etc., and $\mathbb{R}^m$ is the objective space.

\begin{figure}[t]
	\centering
	\begin{subfigure}[b]{.24\textwidth}
		\centering
		\includegraphics[width=0.95\textwidth, trim=1.9cm 1.5cm 2.6cm 1.4cm, clip=true]{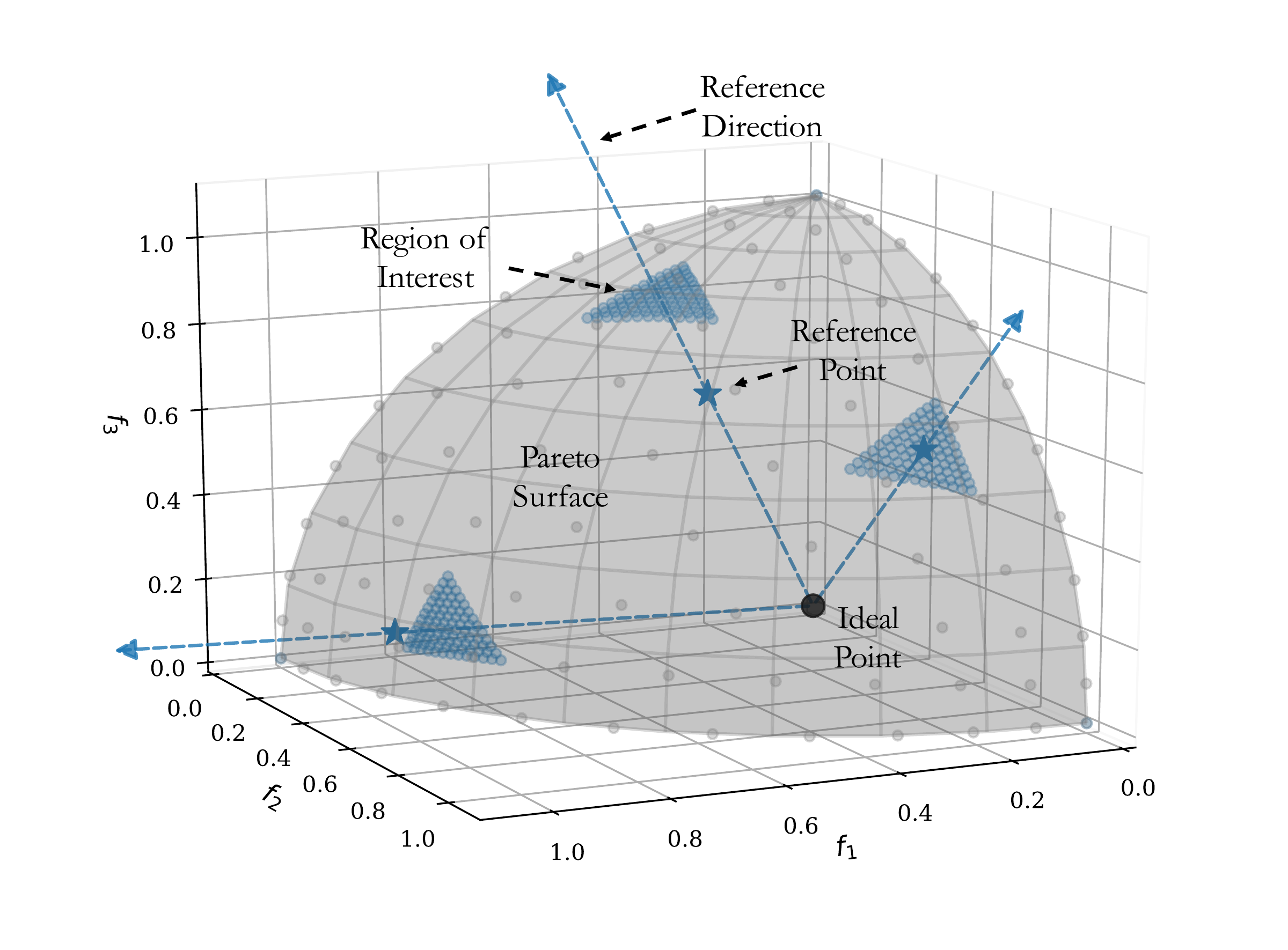}
		\caption{\label{fig:tri_obj}}
	\end{subfigure}\begin{subfigure}[b]{.24\textwidth}
		\centering
		\includegraphics[width=0.95\textwidth, trim=0 0 2cm 0, clip=true]{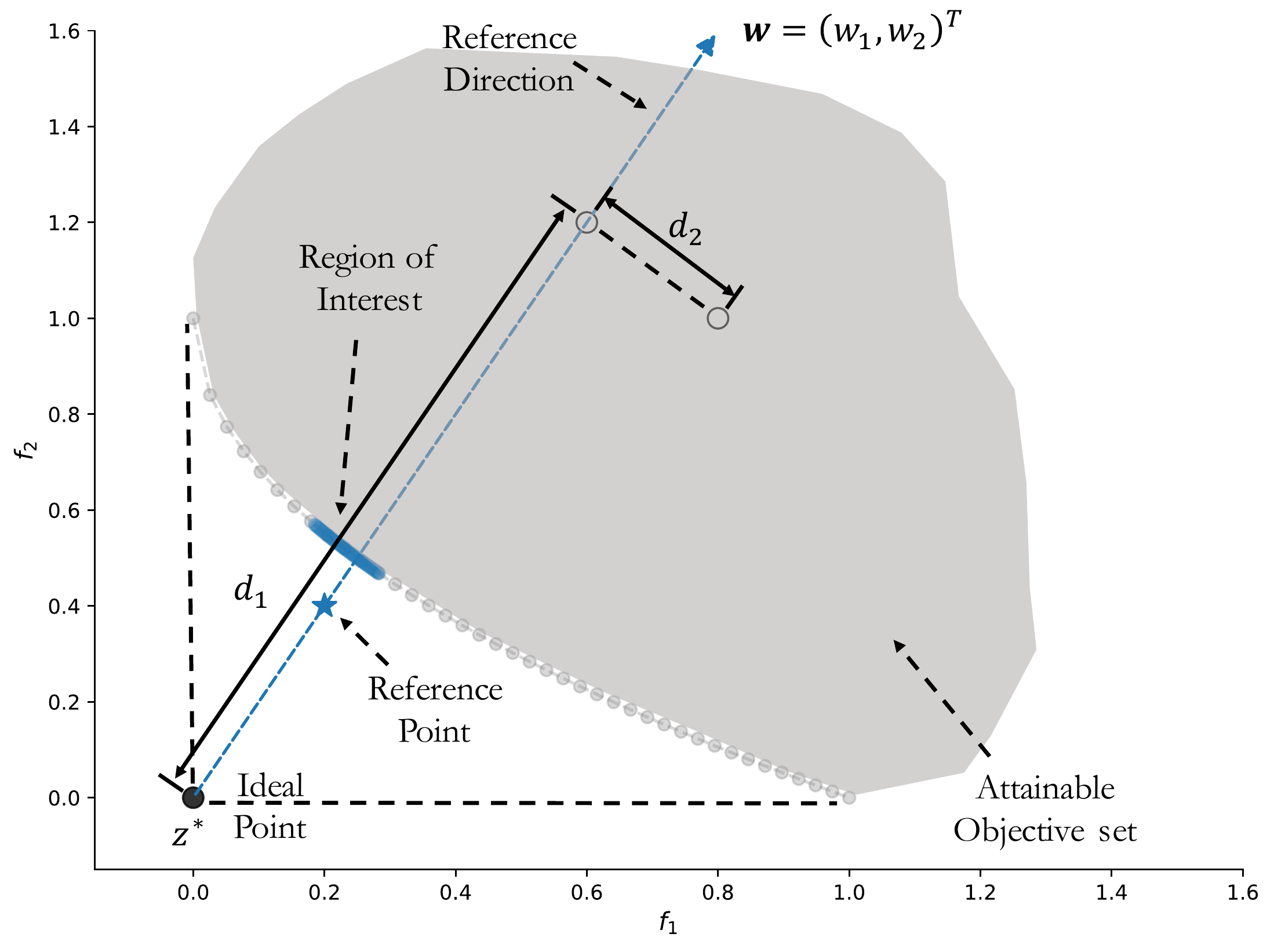}
		\caption{\label{fig:pbi}}
	\end{subfigure}
	\caption{\textbf{Tri-Objective Search:} (a) We leverage user-defined preferences to decompose the tri-objective problem into multiple single-objective sub-problems. By focusing on sub-regions as opposed to the entire Pareto surface, our approach is more efficient. (b) The reference direction is formed by joining the ideal point and user supplied reference targets. The PBI method is used to scalarize the objectives based on the projected distance $d_2$ to the reference target $\bm{w}$, and the distance $d_1$ to the ideal point.\vspace{-0.2cm}}
\end{figure}

As the number of objectives increases, the number of solutions needed to approximate the entire Pareto surface grows exponentially~\cite{deb-book-01}, rendering a global search impractical in most cases. To overcome this challenge we propose a reference guided hyperparameter search. Instead of spanning the entire search space, we focus the hyperparameter search to a neighborhood around few desired user-defined preferences. An illustration of this concept is shown in Fig. \ref{fig:tri_obj}. For instance, in our context, this could correspond to different desired accuracy targets and hardware specifications. This idea enables us to decompose the tri-objective problem into multiple single objective sub-problems. We adopt the penalty-based boundary intersection (PBI) method~\cite{moead} to scalarize multiple objectives into a single objective,
\begin{equation}
    \begin{aligned}
    \Minimize_{} & \hspace{3mm} \textsl{g}^{pbi}(\bm{x} | \bm{w}, \bm{z}^*) = d_1 + \theta d_2\\
    \st & \hspace{3mm} \bm{x} \in \bm{\Omega}, \\
    \end{aligned}
    \label{def:mop_pbi}
\end{equation}
where $d_2=\norm[\bigg]{\textbf{F}(\bm{x}) - \bigg(\bm{z}^* + d_1\frac{\bm{w}}{||\bm{w}||}\bigg)}$, $d_1=\frac{||(\textbf{F}(\bm{x}) - \bm{z}^*)^T \bm{w}||}{||\bm{w}||}$, $\bm{z}^* = (z^*_1, \ldots, z_m^*)^T$ is the ideal objective vector with $z_i^* < \min_{\bm{x}\in\bm{\Omega}}f_i(\bm{x})~i \in \{1, \ldots, m\}$. $\theta \geq 0$ is a trade-off hyperparameter that is set to 5 and $\bm{w}$ is the reference direction obtained by connecting the ideal solution to the desired reference target.

Conceptually, the PBI method constructs a composite measure of the convergence ($d_1$) of the solution to the given reference targets and diversity ($d_2$) of the solutions itself. See Fig.\ref{fig:pbi} for an illustration. In our context, $d_1$ (distance between current projected solution and ideal solution) seeks to push the solution to the boundary of attainable objective space and $d_2$ measures how close the solution is to the user's preference. Finally, we adopt a multi-objective evolutionary algorithm based on decomposition (MOEA/D~\cite{moead}), to simultaneously solve the decomposed sub-problems while optimizing the scalarized objective.

\section{Experiments}
We evaluate the efficacy of \ourmethod{}s on three tasks; image classification, object detection, and transfer learning.
\begin{table*}[!t]
\caption{\textbf{ImageNet Classification~\cite{imagenet}:} \ourmethod{} comparison with manual and automated design of efficient convolutional neural networks. Models are grouped into sections for better visualization. Our results are underlined and the best result in each section is in bold. CPU latency (batchsize=1) is measured on Intel i7-8700K and GPU latency (batchsize=64) is measured on 1080Ti. $^\ddagger$ indicates the objective (in addition to predictive performance) that the method explicitly optimizes through NAS.
\label{tab:imagenet}}
\centering
\resizebox{0.93\textwidth}{!}{%
\scriptsize{
\begin{tabular}{@{\hspace{2mm}}l|c|cc|cc|cc|cc@{\hspace{2mm}}}
\specialrule{1.5pt}{1pt}{1pt}
Model & Type & \#MAdds & Ratio & \#Params & Ratio & CPU(ms)\hspace{-2mm} & GPU(ms) & Top-1 (\%) & Top-5 (\%) \\
\specialrule{1.5pt}{1pt}{1pt}
\textbf{\ourmethod{}-xs (ours)} & auto & \textbf{\underline{66M}}$^\ddagger$ & \underline{1.0x} & \textbf{\underline{1.8M}}$^\ddagger$ & \underline{1.0x} & \underline{6.8} & \underline{18} & \underline{66.7} & \underline{86.8} \\
MobileNetV2\_0.5~\cite{mobilenetv2} & manual & 97M & 1.5x & 2.0M & 1.1x & \textbf{6.2} & 17 & 65.4 & 86.4 \\
MobileNetV3 small~\cite{mobilenetv3} & combined & \textbf{66M} & 1.0x & 2.9M & 1.6x & \textbf{6.2}$^\ddagger$ & \textbf{14} & \textbf{67.4} & - \\
\midrule
\textbf{\ourmethod{}-s (ours)} & auto & \textbf{\underline{117M}}$^\ddagger$ & \underline{1.0x} & \textbf{\underline{2.4M}}$^\ddagger$ & \underline{1.0x} & \underline{9.5} & \underline{25} & \textbf{\underline{71.6}} & \underline{90.3} \\
MobileNetV1~\cite{mobilenet} & manual & 575M & 4.9x & 4.2M & 1.8x & 7.3 & 20 & 70.6 & 89.5 \\
ShuffleNetV2~\cite{shufflenetv2} & manual & 146M & 1.3x & - & - & \textbf{6.8} & \textbf{11}$^\ddagger$ & 69.4 & - \\
ChamNet-C~\cite{chamnet} & auto & 212M & 1.8x & 3.4M & 1.4x & - & - & \textbf{71.6} & - \\
\midrule
\textbf{\ourmethod{}-m (ours)} & auto & \textbf{\underline{218M}}$^\ddagger$ & \underline{1.0x} & \textbf{\underline{3.4M}}$^\ddagger$ & \underline{1.0x} & \underline{14.7} & \underline{42} & \textbf{\underline{75.3}} & \underline{92.5} \\
MobileNetV2~\cite{mobilenetv2} & manual & 300M & 1.4x & 3.4M & 1.0x & \textbf{8.3}$^\ddagger$ & 23 & 72.0 & 91.0 \\
ShuffleNetV2 $2\times$~\cite{shufflenetv2} & manual & 591M & 2.7x & 7.4M & 2.2x & 11.0 & \textbf{22}$^\ddagger$ & 74.9 & -\\
MnasNet-A1~\cite{mnasnet} & auto & 312M & 1.4x & 3.9M & 1.1x & 9.3$^\ddagger$ & 32 & 75.2 & 92.5 \\
MobileNetV3 large~\cite{mobilenetv3} & combined & 219M & 1.0x & 5.4M & 1.6x & 10.0$^\ddagger$ & 33 & 75.2 & - \\
\midrule
\textbf{\ourmethod{}-l (ours)} & auto & \textbf{\underline{318M}}$^\ddagger$ & \underline{1.0x} & \textbf{\underline{4.0M}}$^\ddagger$ & \underline{1.0x} & \underline{19.2} & \underline{74} & \underline{76.6} & \underline{93.2}  \\
MnasNet-A2~\cite{mnasnet} & auto & 340M & 1.1x & 4.8M & 1.2x & - & - & 75.6 & 92.7 \\
FBNet-C~\cite{fbnet} & auto & 375M & 1.2x & 5.5M & 1.4x & \textbf{9.1}$^\ddagger$ & \textbf{31} & 74.9 & - \\
EfficientNet-B0~\cite{efficientnet} & auto & 390M$^\ddagger$ & 1.2x & 5.3M & 1.3x & 14.4 & 46 & 76.3 & 93.2 \\
MixNet-M~\cite{mixnet} & auto & 360M$^\ddagger$ & 1.1x & 5.0M & 1.2x & 24.3 & 79 & \textbf{77.0} & 93.3  \\
\specialrule{1.5pt}{1pt}{1pt}
\end{tabular}}
}
\vspace{-0.2cm}
\end{table*}
\begin{figure*}[t]
	\centering
	\begin{subfigure}[t]{.95\textwidth}
		\centering
		\includegraphics[width=0.98\textwidth]{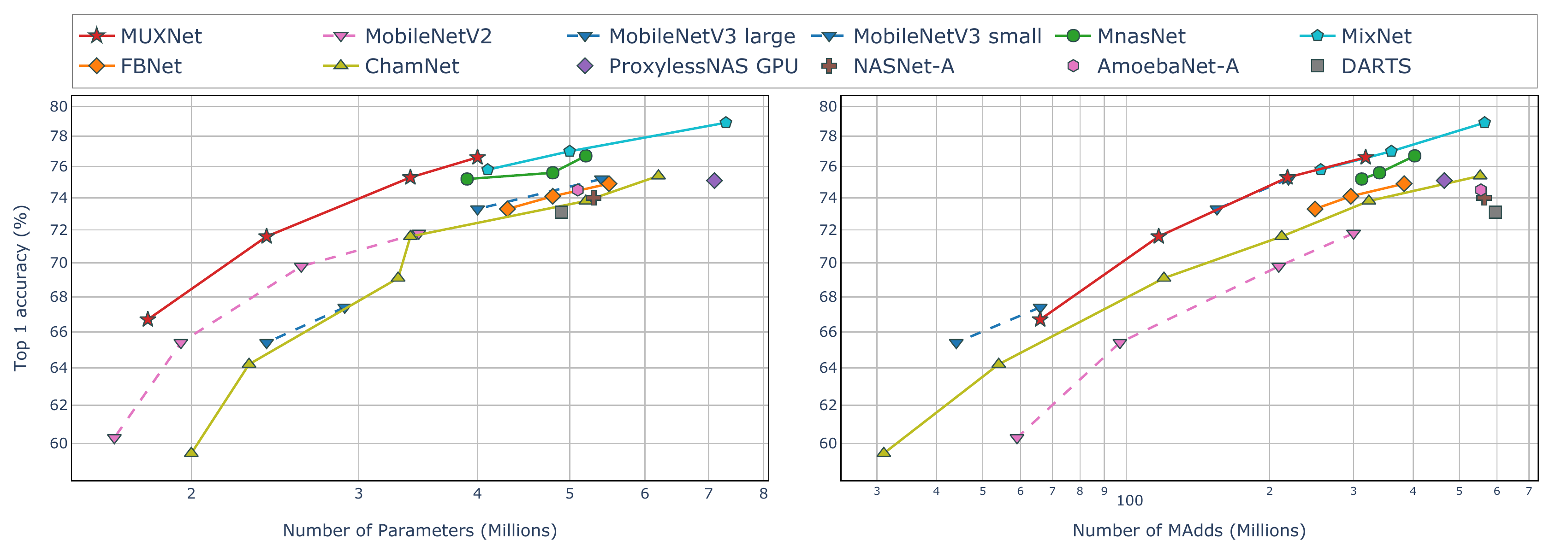}
	\end{subfigure}
	\caption{The trade-off between model complexity and top-1 accuracy on ImageNet. This allows us to compare models designed for different computation requirements in number of parameters or number of multi-adds. All our models use input resolution of $224\times 224$. We use dash line to denote models from channel width multipliers or with different input resolutions.
	\label{fig:imagenet}\vspace{-0.2cm}}
\end{figure*}
\subsection{Hyperparameter Search Details \label{sec:implementation}}
\noindent\textbf{Search Space:} To compensate for the extra hyperparameters introduced by spatial and channel multiplexing, we constrain the commonly adopted layer-wise search space~\cite{proxylessnas,mnasnet,mobilenetv3} to a stage-wise search space, where layers within the same stage share the same hyperparameters. \ourmethod{}s consist of four stages, where each stage begins with a reduction block and is followed by a series of normal blocks. In each stage, we search for kernel size, expansion ratio, repetitions of normal blocks, leave-out ratio for channel multiplexing and the spatial multiplexing settings (see Fig.~\ref{fig:search_space} in Appendix~\ref{sec:search-space}). To further reduce the search space, we always adopt squeeze-and-excitation~\cite{squeezenet} and use swish~\cite{swish} non-linearity for activation at each stage except the first stage, where a ReLU is used.

\noindent\textbf{Search:} Following previous work~\cite{proxylessnas,mnasnet}, we conduct the search directly on ImageNet and estimate model accuracy on a subset consisting of 50K randomly sampled images from the training set. As a common practice, during search, the number of training epochs are reduced to 5. We select four reference points with preferences on model size ranging from 1.5M to 5M, MAdds ranging from 60M to 300M, and predictive accuracy fixed at 1. The compactness and efficiency objectives are normalized between [0, 1] before aggregation. Search is initialized with a global population size of 40 and evolved for 100 iterations, which takes about 11 days on sixteen 2080Ti GPUs. At the end of evolution, we pick the top 5 (based on PBI aggregated function values) models from each of the four subproblems, and retrain them thoroughly from scratch on ImageNet. The four resulting models are named as \ourmethod{}-xs/s/m/l. Architectural details can be found in Appendix~\ref{sec:search-space} (Fig.~\ref{fig:muxnets}).

\subsection{ImageNet Classification}
For training on ImageNet, we follow the procedure outlined in~\cite{mnasnet}. Specifically, we adopt Inception pre-processing with image size \nobyno{224}~\cite{inception-v4}, batch size of 256, RMSProp optimizer with decay 0.9, momentum 0.9, and weight decay 1e-5. A Dropout layer of rate 0.2 is added before the last linear layer. Learning rate is linearly increased to 0.016 in the initial 5 epochs~\cite{lr-warm-up}, it then decays every 3 epochs at a rate of 0.03. We further complement the training with exponential moving average with decay rate of 0.9998.

Table~\ref{tab:imagenet} shows the performance of baselines and MUXNets on ImageNet 2012 benchmark~\cite{imagenet}. We compare them in terms of accuracy on validation set, model compactness (parameter size), model efficiency (MAdds) and inference latency on CPU and GPU. Overall, \ourmethod{}s consistently either match or outperform other models across different accuracy levels. In particular, \ourmethod{}-m achieves 75.3\% accuracy with 3.4M parameters and 218M MAdds, which is \textbf{1.4}$\times$ more efficient and \textbf{1.6}$\times$ more compact when compared to MnasNet-A1~\cite{mnasnet} and MobileNetV3~\cite{mobilenetv3}, respectively. Figures~\ref{fig:preview_results} and \ref{fig:imagenet} visualize the trade-off obtained by \ourmethod{} and previous models. In terms of accuracy and compactness, \ourmethod{} clearly dominates all previous models including MnasNet~\cite{mnasnet}, FBNet~\cite{fbnet}, MobileNetV3~\cite{mobilenetv3}, and MixNet~\cite{mixnet}. In terms of accuracy and efficiency, \ourmethod{}s are on par with current state-of-the-art models, i.e. MobileNetV3 and MixNet.

In terms of latency, the performance of MUXNet models is mixed since they, (i) use non-standard primitives that do not have readily available efficient low-level implementations, and (ii) are not explicitly optimized for latency. Compared to methods that use optimized convolutional primitives but do not directly optimize for latency (EfficientNet/MixNet), MUXNet's latency is competitive despite using unoptimized spatial and channel multiplexing primitives. MUXNet's limitations due to unoptimized implementation can be offset, to an extent, by its inherent FLOPs and parameter efficiency. MUXNet is not as competitive as methods that directly use CPU or GPU latency on Pixel phones as a search objective (MobileNetV3, MnasNet).
\begin{figure*}[t]
	\centering
	\begin{subfigure}[t]{.95\textwidth}
		\centering
		\includegraphics[width=0.98\textwidth]{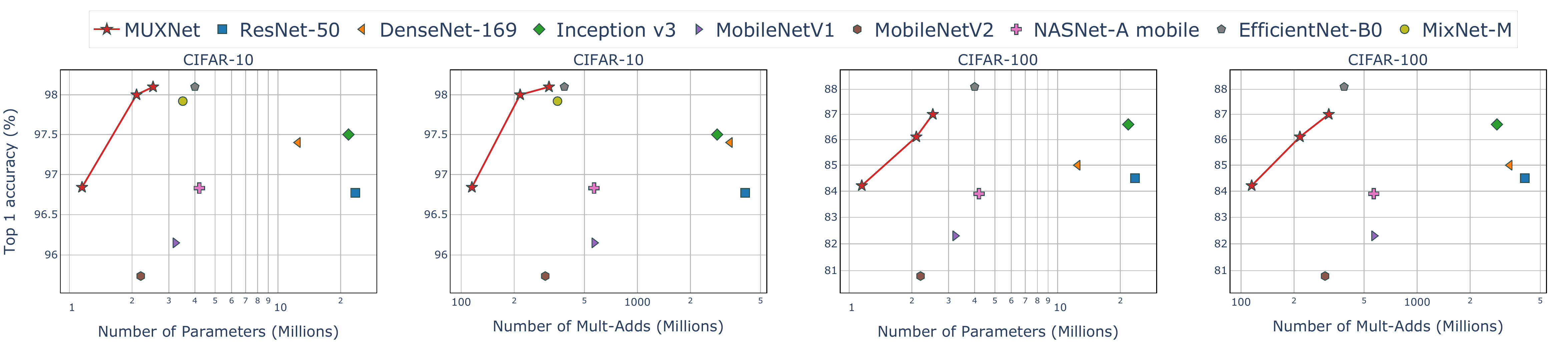}
	\end{subfigure}
	\caption{\textbf{Transfer Learning on CIFAR:} Trade-off between Top-1 accuracy and \#Params / \#MAdds.\label{fig:transfer_cifar}\vspace{-0.2cm}}
\end{figure*}

\subsection{Object Detection}
\begin{table}[h]
\centering
\caption{PASCAL VOC2007~\cite{voc2007} Detection\label{tab:voc2007}}
\resizebox{0.45\textwidth}{!}{%
\begin{tabular}{@{\hspace{2mm}}l|cc|c@{\hspace{2mm}}}
\specialrule{1.5pt}{1pt}{1pt}
Network & \#MAdds & \#Params & mAP (\%) \\ \midrule
VGG16 + SSD~\cite{ssd} & 35B & 26.3M & 74.3 \\
MobileNet + SSD~\cite{huang2017speed} & 1.6B & 9.5M & 67.6 \\
MobileNetV2 + SSDLite~\cite{mobilenetv2} & 0.7B & 3.4M & 67.4 \\
MobileNetV2 + SSD~\cite{mobilenetv2} & 1.4B & 8.9M & 73.2 \\ \midrule
\textbf{\ourmethod{}-m + SSDLite (ours)} & \textbf{0.5B} & \textbf{3.2M} & \textbf{68.6} \\
\textbf{\ourmethod{}-l + SSD (ours)} & \textbf{1.4B} & \textbf{9.9M} & \textbf{73.8} \\
\specialrule{1.5pt}{1pt}{1pt}
\end{tabular}}
\vspace{-0.2cm}
\end{table}
We evaluate and compare the generalization ability of \ourmethod{} and other peer models on the PASCAL VOC detection benchmark~\cite{voc2007}. Our experiments use both the Single Shot Detector (SSD)~\cite{ssd} and the Single Shot Detector Lite (SSDLite)~\cite{mobilenetv2} as the detection frameworks, with \ourmethod{} as the feature extraction backbone. We follow the procedure in~\cite{mobilenetv2} to setup the additional prediction layers, i.e. location of detection heads in the backbone, size of corresponding boxes, etc. The combined \emph{trainval} sets of PASCAL VOC 2007 and 2012 are used for training. Other details include, SGD optimizer with momentum 0.9 and weight decay 5e-4, batch size of 32, input image resized to \nobyno{300} and learning rate of 0.01 with cosine annealing to 0.0 in 200 epochs. Table~\ref{tab:voc2007} reports the mean Average Precision (mAP) on the PASCAL VOC 2007 test set. When paired with the same detector framework SSDLite, our \ourmethod{}-m model achieves 1.2\% higher mAP than MobileNetV2~\cite{mobilenetv2} while being 6\% more compact and $1.4\times$ more efficient.
\subsection{Transfer Learning}
To further explore the efficacy of \ourmethod{} we evaluate it under the transfer learning setup in~\cite{transfer_learning} on three different datasets; CIFAR-10, CIFAR-100 and ChestX-Ray14~\cite{wang2017chestx}.
\subsubsection{CIFAR-10 and CIFAR-100}
Both CIFAR-10 and -100 datasets have 50,000 and 10,000 images for training and testing, respectively. CIFAR-100 extends CIFAR-10 by adding 90 more classes resulting in 10$\times$ fewer training examples per class. For training on both datasets, the models are initialized with weights pre-trained on ImageNet. The model is then fine-tuned using SGD with momemtum 0.9, weight decay 4e-5 and gradients clipped to a magnitude of 5. Learning rate is set to 0.01 with cosine annealing to 0.0 in 150 epochs. For data augmentation, images are up-sampled via bicubic interpolation to \nobyno{224} and horizontally fliped at random. Table~\ref{tab:cifars} and Figure~\ref{fig:transfer_cifar} reports the accuracy, compactness and efficiency of \ourmethod{} and other baselines. Overall, \ourmethod{} significantly outperforms previous methods on both CIFAR-10 and -100 datasets, indicating that our models also transfer well to other similar tasks. In particular, \ourmethod{}-m achieves 1\% higher accuracy than NASNet-A mobile with \textbf{3}$\times$ fewer parameters while being \textbf{2}$\times$ more efficient in MAdds.
\subsubsection{ChestX-Ray14}
The ChestX-Ray14 benchmark was recently introduced in~\cite{wang2017chestx}. The dataset consists of 112,120 high resolution frontal-view chest X-ray images from 30,805 patients. Each image is labeled with one or multiple common thorax diseases, or ``Normal'', otherwise. Due to the multi-label nature of the dataset, we use a multitask learning setup where each disease is treated as an individual binary classification problem. We define a 14-dimensional label vector of binary values indicating the presence of one or more diseases, and optimize a regression loss as opposed to cross-entropy in single-label cases. The training procedure is similar to the CIFAR experiments for transfering pre-trained models. Table~\ref{tab:chexray} compares the performance of \ourmethod{}-m with previous approaches, including CheXNet~\cite{rajpurkar2017chexnet} which represents the state-of-the-art on this dataset. Evidently, \ourmethod{}-m's performance in terms of area under the receiver operating characteristic (AUROC) curve on the test set is comparable (84.1\% vs 84.4\%) to CheXNet while being \textbf{3}$\times$ more compact and \textbf{14}$\times$ more efficient.
\begin{table}[t]
\centering
\caption{\textbf{Transfer Learning:} Top-1 accuracy on CIFAR-10 (C-10) and CIFAR-100 (C-100). ResNet, DenseNet, MobileNetV2, and NASNet-A results are from~\cite{transfer_learning}.\label{tab:cifars}}
\resizebox{0.45\textwidth}{!}{%
\begin{tabular}{@{\hspace{2mm}}l|cc|cc@{\hspace{2mm}}}
\specialrule{1.5pt}{1pt}{1pt}
Model & \#MAdds & \#Params & C-10 (\%) & C-100 (\%) \\ \midrule
ResNet-50~\cite{resnet} & 4.1B &  23.5M & 96.77 & 84.50 \\
DenseNet-169~\cite{densenet} & 3.4B & 12.5M & 97.40 & 85.00 \\
MobileNetV2~\cite{mobilenetv2} & 0.3B & 2.2M & 95.74 & 80.80 \\
NASNet-A mobile~\cite{nasnet} & 0.6B & 4.2M & 96.83 & 83.90 \\
EfficientNet-B0~\cite{efficientnet} & 0.4B & 4.0M & 98.10 & 88.10 \\
MixNet-M~\cite{mixnet} & 0.4B & 3.5M & 97.92 & - \\ \midrule
\textbf{\ourmethod{}-m (ours)} & \textbf{0.2B} & \textbf{2.1M} & \textbf{98.00} & \textbf{86.11} \\
\specialrule{1.5pt}{1pt}{1pt}
\end{tabular}%
}
\vspace{-0.2cm}
\end{table}
\begin{table}[t]
\centering
\caption{Transfer Learning on ChestX-Ray14~\cite{wang2017chestx}\label{tab:chexray}}
\resizebox{0.45\textwidth}{!}{
\begin{tabular}{@{\hspace{2mm}}l|cc|c@{\hspace{2mm}}}
\specialrule{1.5pt}{1pt}{1pt}
Method  & \#MAdds  & \#Params & Test AUROC (\%) \\ \midrule
Wang et al. (2017)~\cite{wang2017chestx}  & - & -        & 73.8            \\
Yao et al. (2017) ~\cite{yao2017learning}  & - & -        & 79.8            \\
CheXNet (2017)~\cite{rajpurkar2017chexnet}  & 2.8B & 7.0M    & 84.4            \\
\midrule
\textbf{\ourmethod{}-m (ours)} & \textbf{0.2B}  & \textbf{2.1M}    & \textbf{84.1}            \\
\specialrule{1.5pt}{1pt}{1pt}
\end{tabular}
}
\vspace{-0.3cm}
\end{table}


\section{Ablation Study\label{sec:ablation}}
\begin{figure*}[t]
	\centering
	\begin{subfigure}[b]{.35\textwidth}
		\centering
		\includegraphics[width=0.98\textwidth]{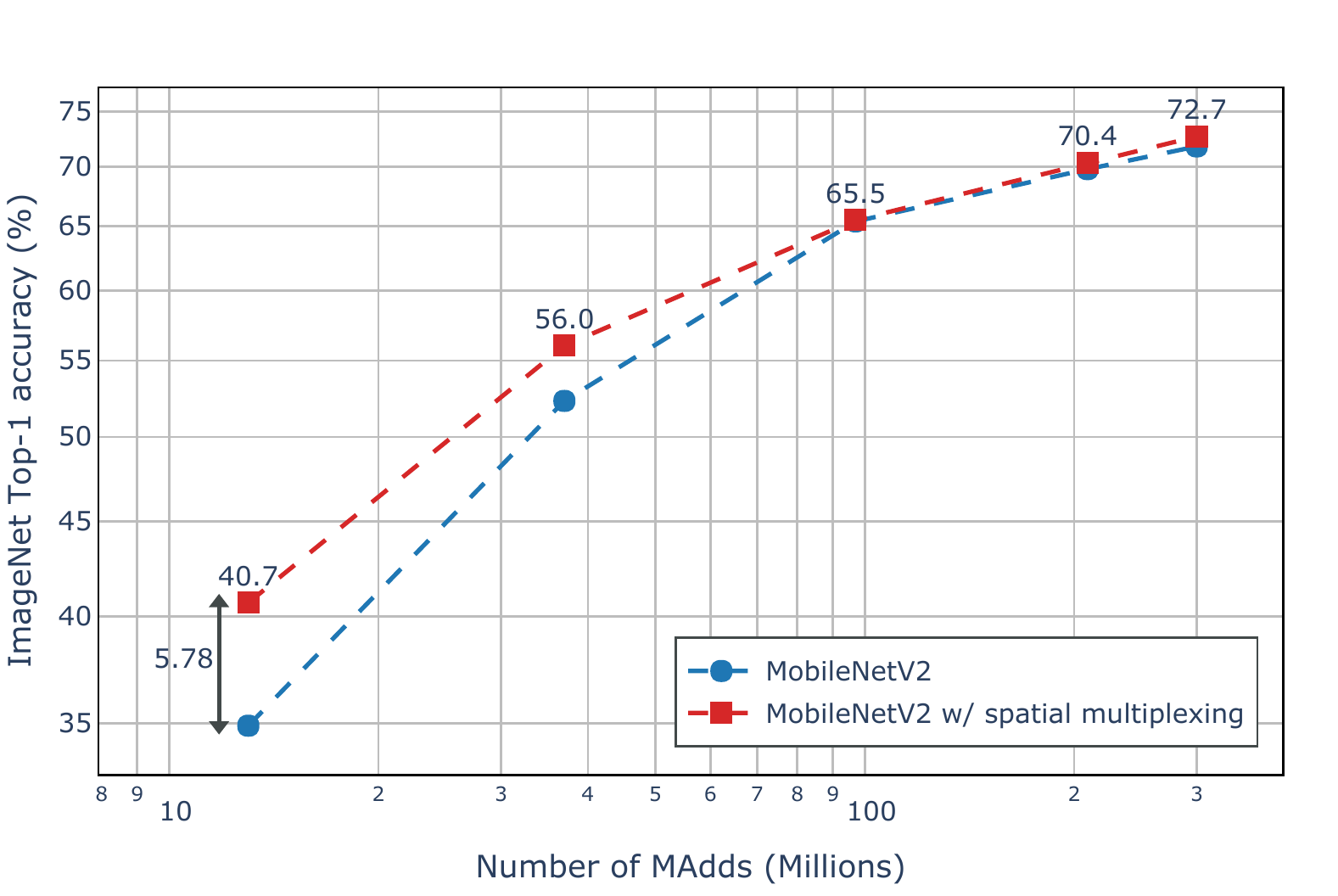}
		\caption{Spatial Multiplexing\label{fig:ablation_spatial_multiplexing}\vspace{-0.3cm}}
	\end{subfigure}
	\begin{subfigure}[b]{.6\textwidth}
		\centering
		\includegraphics[width=0.98\textwidth]{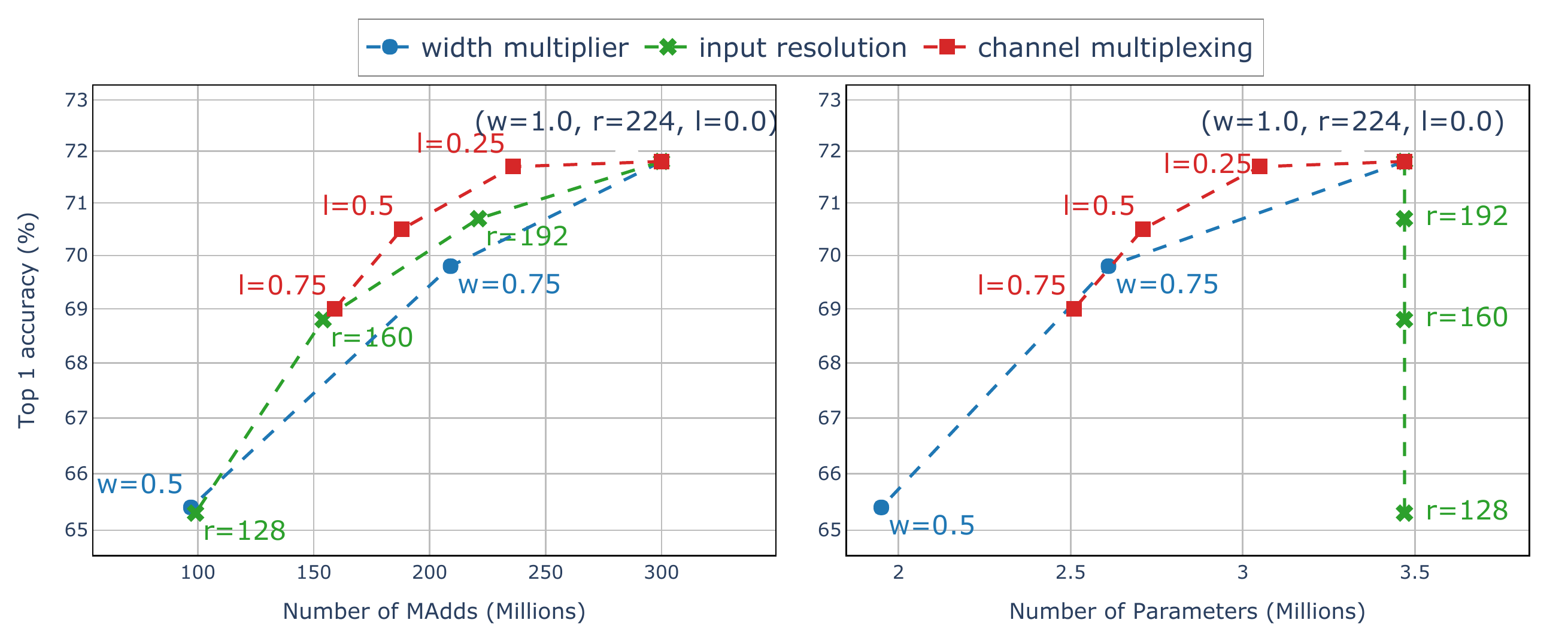}
		\caption{Channel Multiplexing\label{fig:ablation_channel_multiplexing}\vspace{-0.3cm}}
	\end{subfigure}
	\caption{\textbf{Multiplexed Convolution Ablation Study:} (a) Results correspond to width multiplier of $0.1$, $0.25$, $0.5$, $0.75$, and $1.0$. (b) \emph{w}, \emph{r} and \emph{l} are width multiplier, input resolution and leave-out ratio, respectively. When $l=0.25$, 75\% of the input information is processed at each normal block.\vspace{-0.3cm}}
\end{figure*}

\vspace{3pt}
\noindent\textbf{Spatial Multiplexing:} We incorporate the spatial multiplexing operation within the \nobyno{3}\ depth-wise separable convolution layers of MobileNetV2. As we do in our main experiments, we do not apply spatial multiplexing to the reduction blocks. We manually fix the multiplexing hyperparameters to $C_1=C_3=\frac{C}{4}, C_2=\frac{C}{2}$ i.e., 1/4 channels are processed by subpixeling, 1/4 of the channels are processed by superpixeling, and the remaining channels are processed without modification. Figure~\ref{fig:ablation_spatial_multiplexing} shows the effect of spatial multiplexing on MobileNetV2~\cite{mobilenetv2} at different width multipliers. Spatial multiplexing consistently improves accuracy over the original depth-wise separable convolution at fixed spatial resolution. In particular, spatial multiplexing boosts accuracy by \textbf{5.8}\% in low MAdds regime. The results suggest that per MAdd, spatial multiplexing (groups+full conv) has better information flow than dep-sep+$1\times 1$ conv. This is more apparent in small models which have less channels, so $1\times 1$ conv cannot effectively mix channel information.

\vspace{3pt}
\noindent\textbf{Channel Multiplexing:} To make models more efficient, methods such as scaling down the number of channels by a factor (named width multiplier), or scaling down the input resolution have been proposed. Here we investigate the impact of channel multiplexing as an alternative to reduce model complexity. To be consistent with the main experiments we only apply channel multiplexing to the normal blocks. In MobileNetV2~\cite{mobilenetv2} we gradually increase the number of input channels that are left unprocessed in each normal block. We use $l$ to denote the leave-out ratio, where a high value corresponds to less channels being processed and hence more efficiency. The resulting trade-off with accuracy is shown in Figure~\ref{fig:ablation_channel_multiplexing}. Evidently, reducing the resolutions of input images provides a better trade-off between accuracy and MAdds than reducing the channels. However, reducing the input resolution provides no benefit to model size. On the other hand, channel multiplexing offers competitive trade-off in both cases; MAdds and model size. In particular, leaving out 25\% of the input channels at every normal block appears to affect the predictive accuracy minimally, while simultaneously saving \textbf{13}\% in parameters and \textbf{20}\% in multiply-adds.

\begin{figure}[t]
	\centering
	\begin{subfigure}[b]{.45\textwidth}
		\centering
		\includegraphics[width=0.98\textwidth]{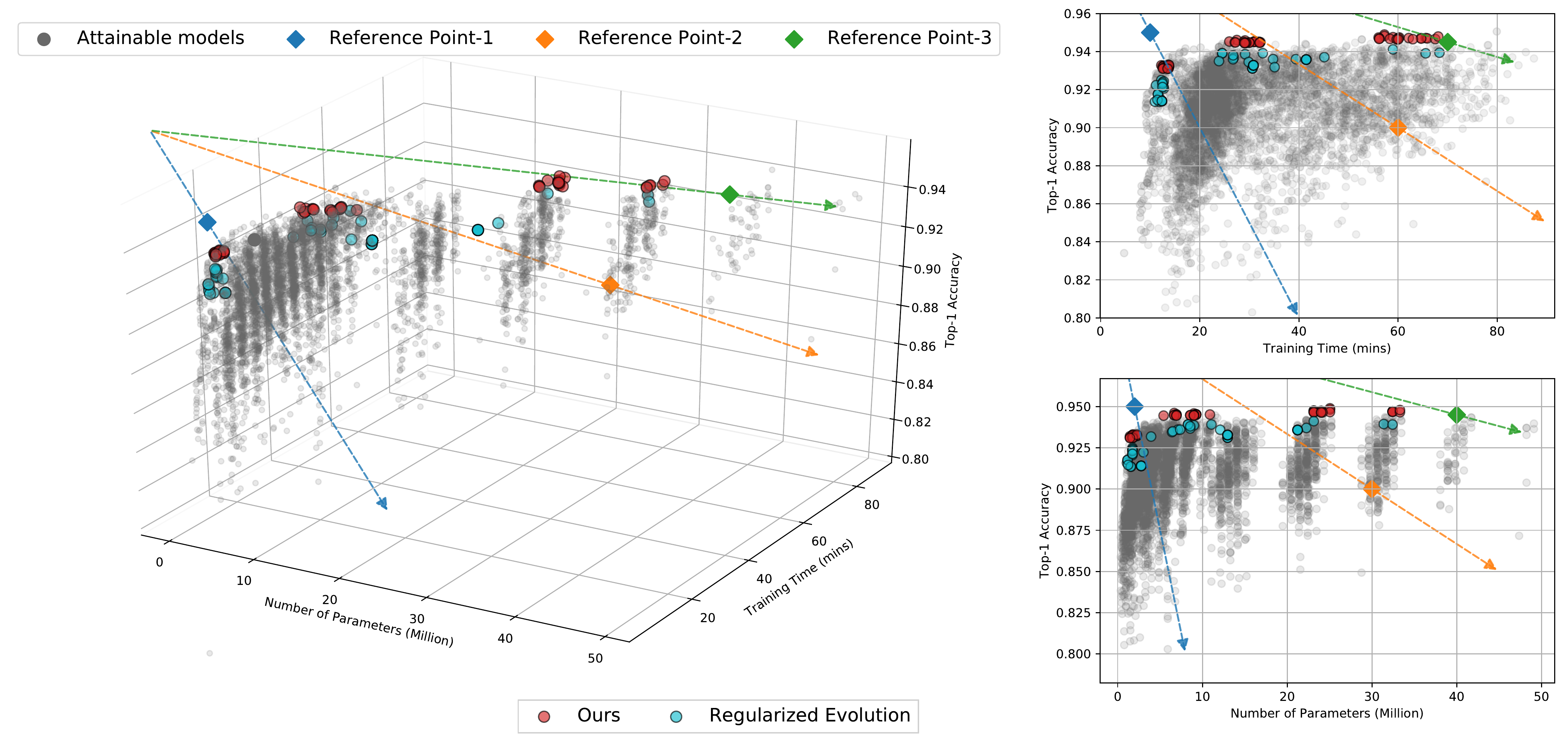}
	\end{subfigure}
	\caption{Performance comparison between our approach and regularized evolution (RE)~\cite{amoebanet} on NASBench101~\cite{nasbench101}. Both methods are subject to the same search budget of 1,000 maximum models sampled. We distribute the search budget across three executions of RE for each one of the three reference points. Our approach simultaneously targets all three reference points in one run using all available budget.
	\label{fig:nasbench}\vspace{-0.3cm}}
\end{figure}

\vspace{3pt}
\noindent\textbf{Search Efficiency:} To thoroughly and efficiently evaluate the effectiveness of the PBI decomposition technique and the search efficiency of our proposed NAS algorithm, we adopt the NASBench101~\cite{nasbench101} benchmark. It contains more than 400K unique models pre-trained on CIFAR-10, whose Pareto-optimal solutions and predictive performance are readily available without expensive training. In this case, we aim to minimize the number of parameters, the training time and maximize the accuracy. We also adopt the regularized evolution~\cite{amoebanet} approach as a baseline for comparison. Figure~\ref{fig:nasbench} shows the search effectiveness for three reference points under a fixed computational budget. The PBI scalarization is effective in directing the search towards pre-defined target regions as the obtained solutions from both methods are centered around the three provided target points. In addition, we observe that by collectively solving the sub-problems, we achieve better results under the same search budget as opposed to solving the sub-problem one at a time, as in case of regularized evolution.


\section{Conclusion}
This paper introduced MUXConv, an efficient alternative to a standard convolutional layer that is designed to progressively multiplex channel and spatial information in the network. Furthermore, we coupled it with an efficient multi-objective evolutionary algorithm based hyperparameter search to trade-off predictive accuracy, model compactness and computational efficiency. Experimental results on image classification, object detection and transfer learning suggest that \ourmethod{}s are able to match the predictive accuracy and efficiency of current state-of-the-art models while be more compact.

\vspace{3pt}
\noindent\textbf{Acknowledgements:} We gratefully acknowledge Dr. Erik Goodman and Dr. Wolfgang Banzhaf for partially supporting the computational requirements of this work. Vishnu Naresh Boddeti was partially supported by the Ford-MSU Alliance.

\begin{appendices}

\begin{figure*}[ht]
	\centering
	\begin{subfigure}[b]{.85\textwidth}
		\centering
		\includegraphics[width=0.98\textwidth]{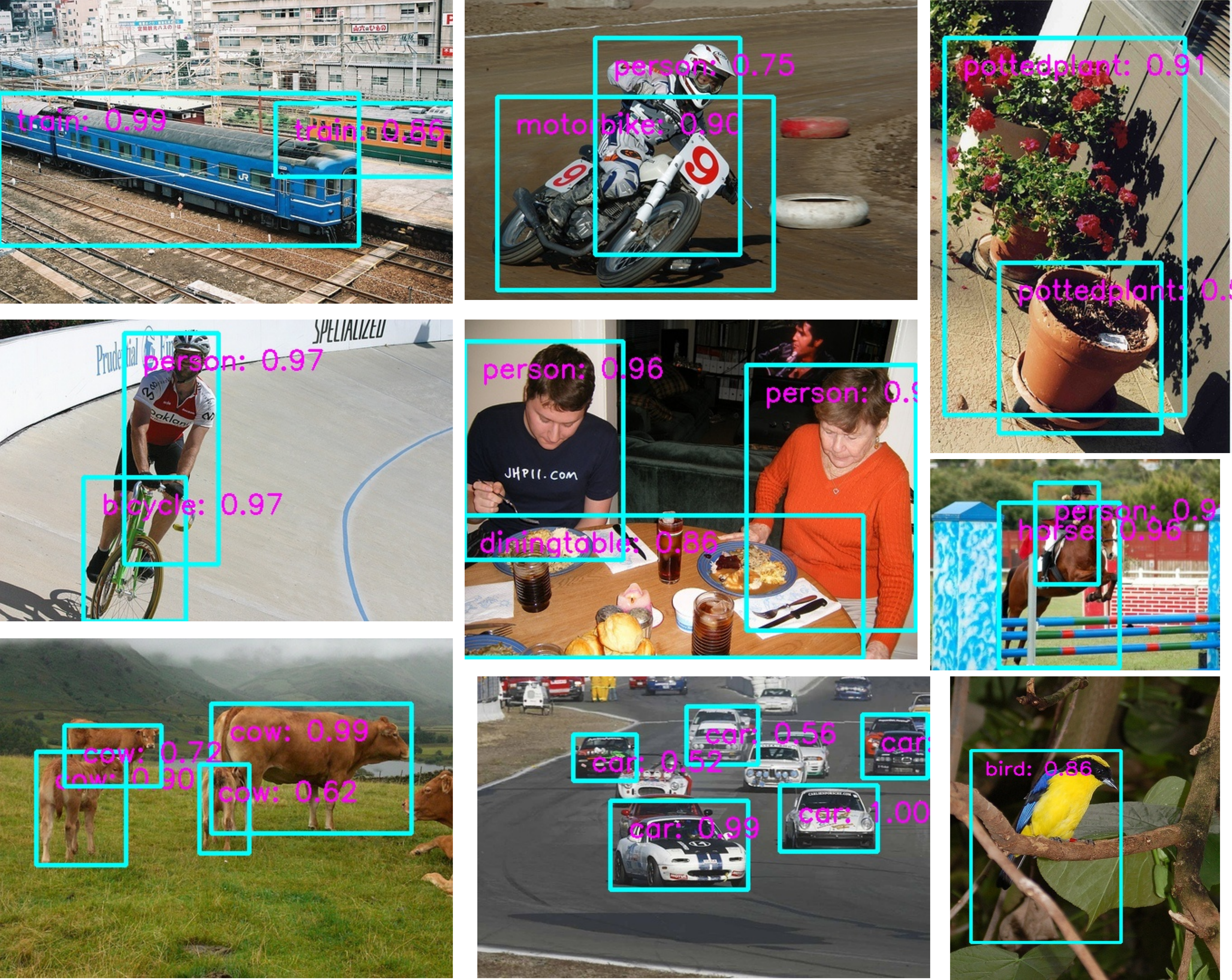}
	\end{subfigure}
	\caption{Examples visualizing the detection performance of \ourmethod{}-m on PASCAL VOC 2007~\cite{voc2007}.
	\label{fig:detection}}
\end{figure*}

\begin{figure*}[ht]
	\centering
	\begin{subfigure}[b]{.85\textwidth}
		\centering
		\includegraphics[width=0.98\textwidth]{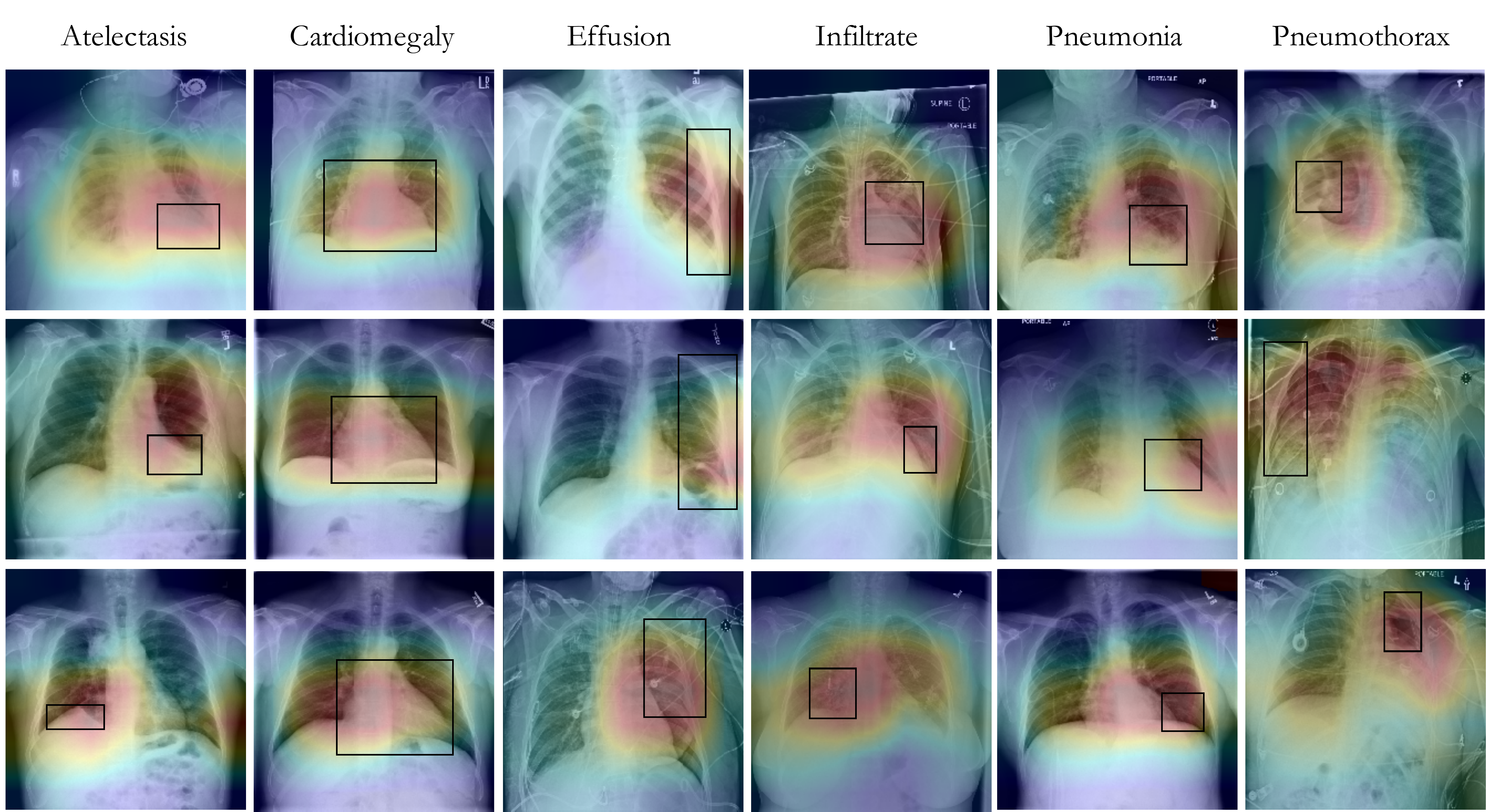}
	\end{subfigure}
	\caption{Examples of class activation map~\cite{cam} of \ourmethod{}-m on ChestX-Ray14~\cite{wang2017chestx}, highlighting the class-specific discriminative regions. The ground truth bounding boxes are plotted over the heatmaps.
	\label{fig:chestXray_gradcam}}
\end{figure*}

In this Appendix we include (1) MUXNet hyperparameter search space in Section \ref{sec:search-space}, (2) computational complexity of MUXNet and comparison to a combination of $1\times 1$ + $3\times 3$ in Section \ref{sec:computational-complexity}, (3) effectiveness of MUXNet as a backbone semantic segmentation in Section \ref{sec:semantic-segmentation}, and (4) evaluation of generalization and robustness properties of MUXNet in Section \ref{sec:generalization}. Finally Fig. \ref{fig:detection} shows some qualitative object detection results on PASCAL VOC 2007, and Fig. \ref{fig:chestXray_gradcam} shows gradCam results on the ChestX-Ray14 dataset.

\section{Search Space \label{sec:search-space}}
\begin{figure}[!tbh]
	\centering
	\begin{subfigure}[t]{.48\textwidth}
		\centering
		\includegraphics[width=0.95\textwidth]{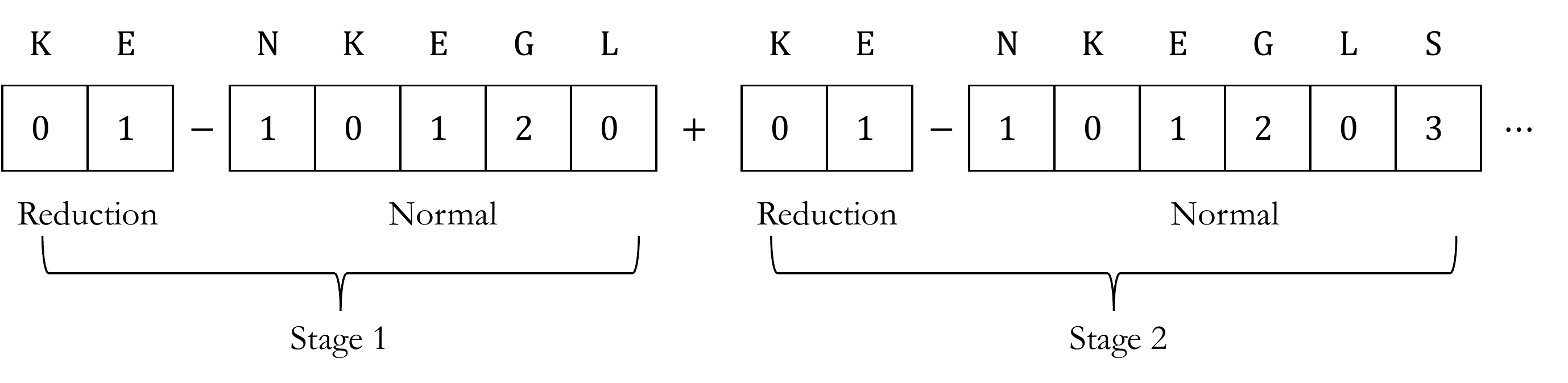}
	\end{subfigure}
	\caption{Search Space Encoding: Each stage is encoded as an integer string. Genetic operations are performed on such encoding. See Table~1 for full details on the options.\label{fig:search_space}\vspace{-0.3cm}}
\end{figure}

\begin{figure*}[t]
	\centering
	\begin{subfigure}[t]{.24\textwidth}
		\centering
		\includegraphics[width=0.95\textwidth,angle=0]{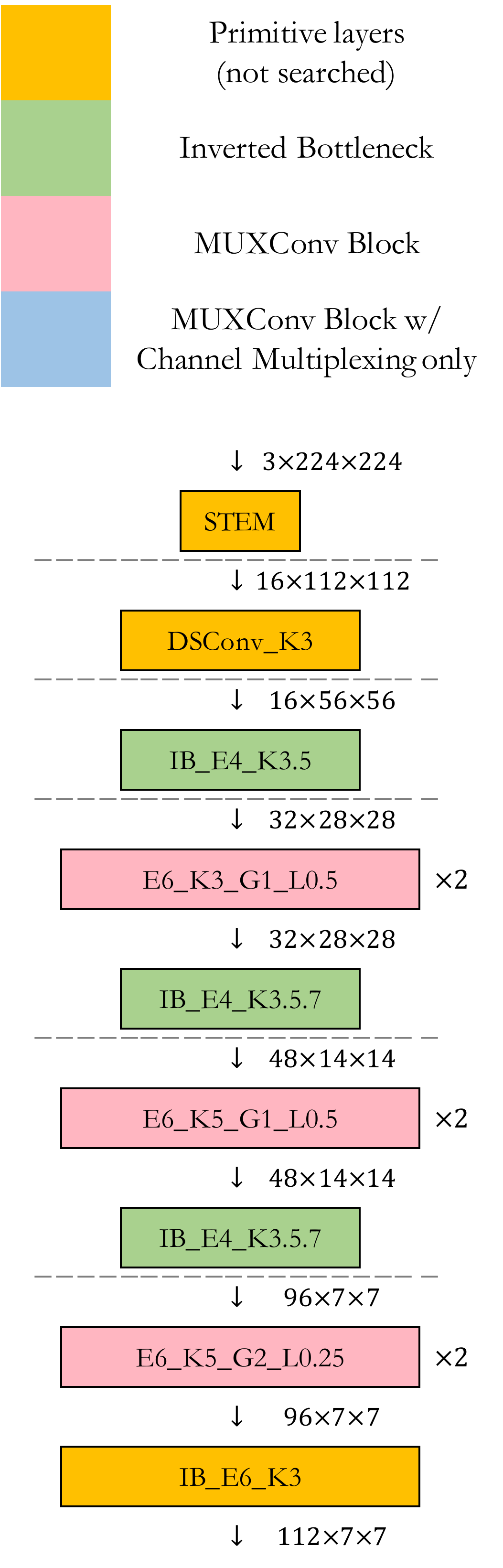}
		\caption{\ourmethod{}-xs\label{fig:muxnet-xs}\vspace{-0.3cm}}
	\end{subfigure}
	\begin{subfigure}[t]{.24\textwidth}
		\centering
		\includegraphics[width=0.95\textwidth,angle=0]{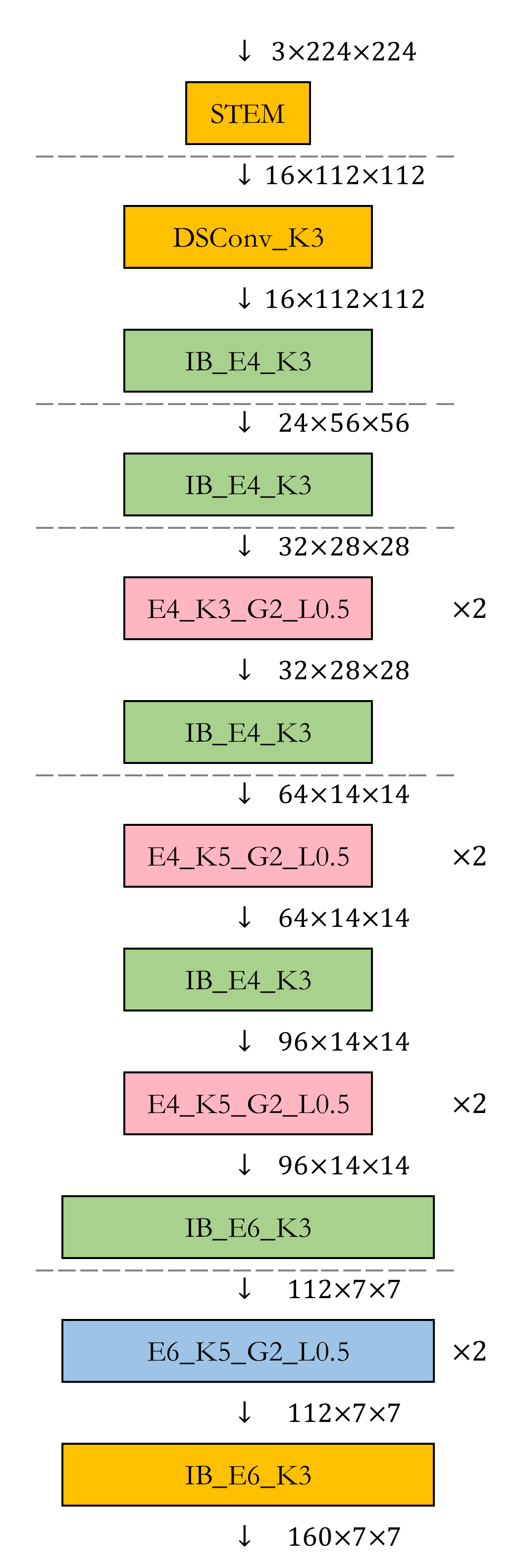}
		\caption{\ourmethod{}-s\label{fig:muxnet-s}\vspace{-0.3cm}}
	\end{subfigure}
	\begin{subfigure}[t]{.24\textwidth}
		\centering
		\includegraphics[width=0.95\textwidth,angle=0]{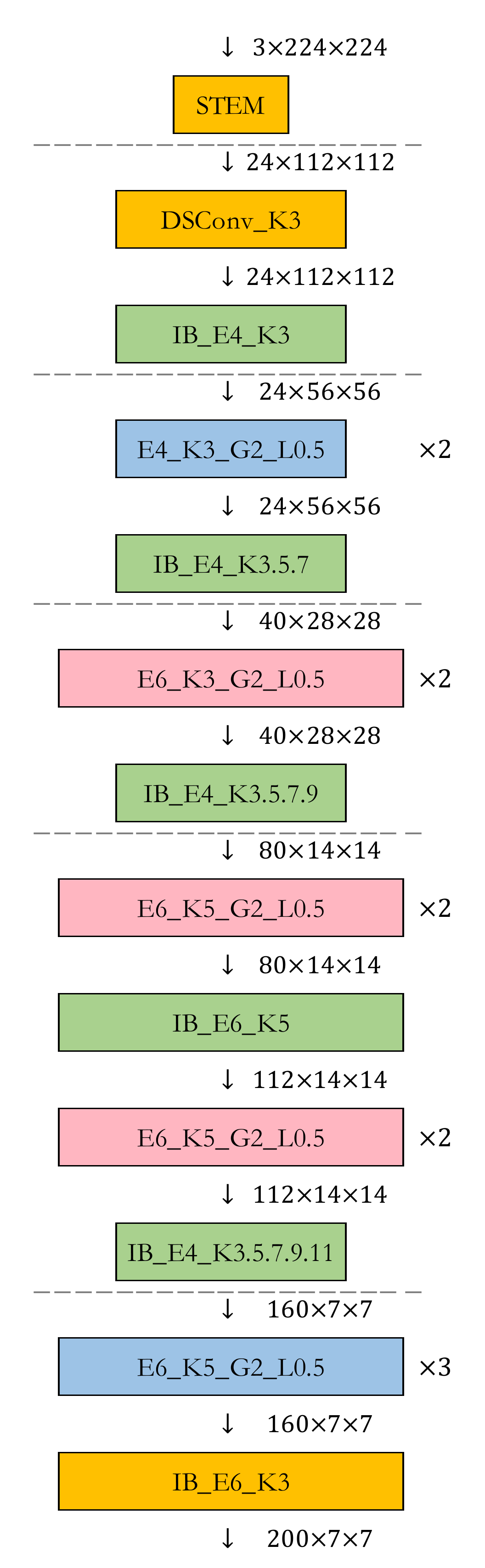}
		\caption{\ourmethod{}-m\label{fig:muxnet-m}\vspace{-0.3cm}}
	\end{subfigure}
	\begin{subfigure}[t]{.24\textwidth}
		\centering
		\includegraphics[width=0.95\textwidth,angle=0]{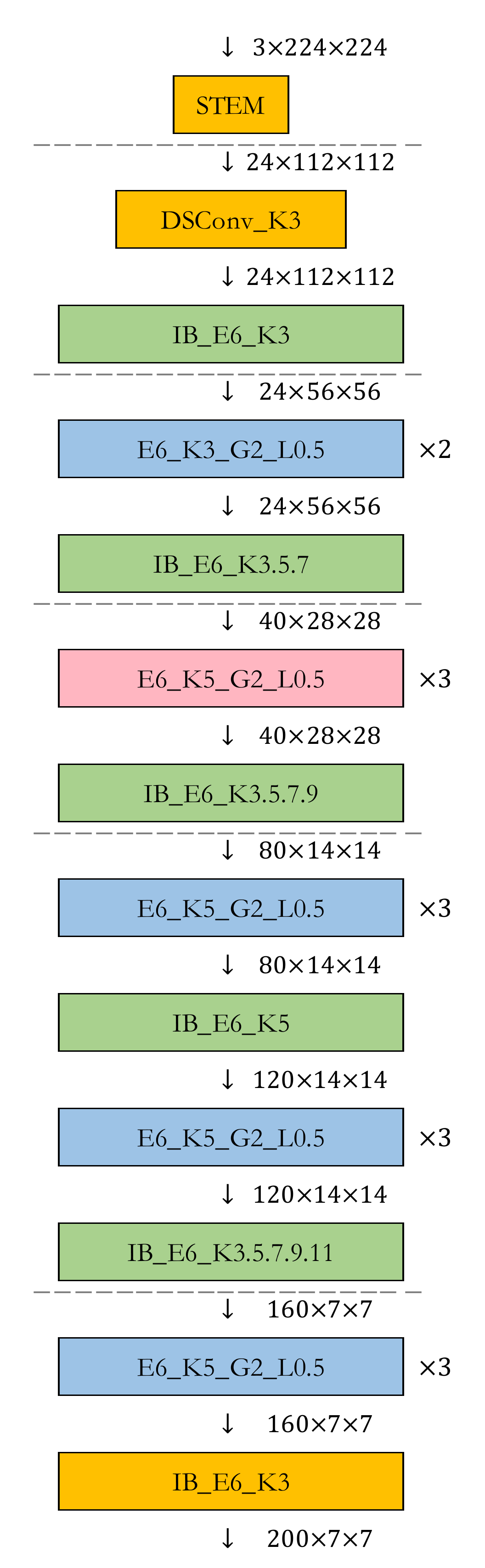}
		\caption{\ourmethod{}-l\label{fig:muxnet-l}\vspace{-0.3cm}}
	\end{subfigure}
	\caption{The architectures of \ourmethod{}-xs/s/m/l in Table~1 (main paper). All architectures share the same hyperparameter settings (except \# of output channels) for the blocks colored in yellow and they are fixed manually. The Dash lines indicate down-sampling points and we divide the architectures into four main stages. We use \emph{E}, \emph{K}, \emph{G}, and \emph{L} to denote expansion rate, kernel size, number of channels per group and leave-out ratio, respectively. Blocks colored in green use the inverted bottleneck structure proposed in~\cite{mobilenetv2}. Blocks colored in pink use both spatial and channel multiplexing and blocks colored in blue only use channel multiplexing.
	\label{fig:muxnets}\vspace{-0.3cm}}
\end{figure*}

To encode the hyperparameter settings for a model, we first divide the model architectures into four stages, based on the spatial resolution of each layer's output feature map. In each stage spatial resolution does not change. The first layer in each stage reduces the feature map size by half. For each stage, we search for kernel size (K) and expansion ratio (E). In addition, from second layer in each stage, we search for \# of repetitions (N), \# of input channels to compute convolution (G), leave-out ratio in channel multiplexing (L) and the spatial multiplexing setting (S) (see Fig.~\ref{fig:search_space}). Table~\ref{tab:search-space} summarizes the hyperparameters and available options for each stage. The obtained hyperparameters for our \ourmethod{}s are visualized in Figure~\ref{fig:muxnets}. The total volume of the search space is approximately $14^{12}$.

\begin{table}[ht]
\centering
\resizebox{0.45\textwidth}{!}{%
\begin{tabular}{@{\hspace{2mm}}l|lc|l|l@{\hspace{2mm}}}
\specialrule{1.5pt}{1pt}{1pt}
 & Hyperparameter & Notation & Options & Stages \\
\specialrule{1.5pt}{1pt}{1pt}
\multirow{6}{*}{\begin{tabular}[c]{@{}l@{}}Normal \\ Blocks\end{tabular}} & Kernel size & K & \{3, 5\} & \{1, 2, 3, 4\} \\
 & Expansion rate & E & \{4, 6\} & \{1, 2, 3, 4\} \\
 & Group factor & G & \{1, 2, 4\} & \{1, 2, 3, 4\} \\
 & Repetitions & N & \{0, 1, 2, 3\} & \{1, 2, 3, 4\} \\
 & Leave-out ratio & L & \{0.0, 0.25, 0.5\} & \{1, 2, 3, 4\} \\
 & \multirow{2}{*}{Spatial Mux} & \multirow{2}{*}{S} & \multirow{2}{*}{\begin{tabular}[c]{@{}l@{}}\{0, {[}-1, 0, 0{]}, {[}0, 0, 1{]},\\{[}1, 0, 1{]}, {[}-1, 0, 0, 1{]}\}\end{tabular}} & \multirow{2}{*}{\{2, 3\}} \\
 &  &  &  &  \\ \midrule
\multirow{2}{*}{\begin{tabular}[c]{@{}l@{}}Reduction\\ Blocks\end{tabular}} & Kernel size & K & \{3, {[}3, 5, 7{]}, {[}3, 5, 7, 9{]}\} & \{1, 2, 3, 4\} \\
 & Expansion rate & E & \{4, 6\} & \{1, 2, 3, 4\} \\
 \specialrule{1.5pt}{1pt}{1pt}
\end{tabular}%
}
\caption{Hyperparameter search space summary. The searched hyperparameters depend on both the block type---i.e. normal or reduction block, and the stages. In case of spatial mutiplexing, option ``-1" means subpixel multiplexing, ``1'' means superpixel multiplexing, and ``0'' means no spatial multiplexing. For instance, ``{[}-1, 0, 1{]}'' means applying subpixel to 1/3 of the input channels, superpixel to another 1/3 of the input channels, and the remaining 1/3 are processed at the original resolution. And we only apply spatial mutiplexing in stages two and three. For the kernel size options in case of reduction blocks, we allow multiple parallel kernels to down-sample the resolution, for example, ``{[}3, 5, 7{]}'' means three parallel convolutions with kernel size of 3, 5, and 7.
\label{tab:search-space}}
\end{table}

\section{Computational Complexity\label{sec:computational-complexity}}
\begin{figure}[h]
	\centering
	\begin{subfigure}[t]{.23\textwidth}
		\centering
		\includegraphics[width=0.9\textwidth]{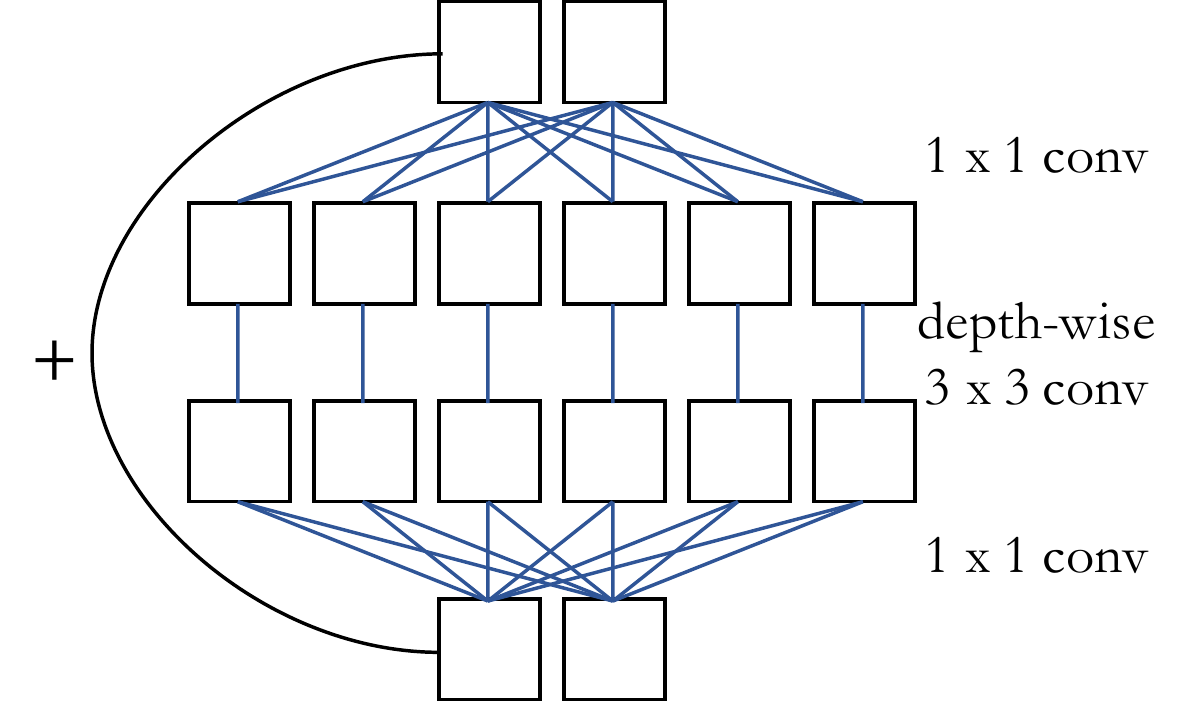}
		\caption{Mobilenet block~\cite{mobilenetv2}\label{fig:irb}\vspace{-0.3cm}}
	\end{subfigure}
	\begin{subfigure}[t]{.23\textwidth}
		\centering
		\includegraphics[width=0.9\textwidth]{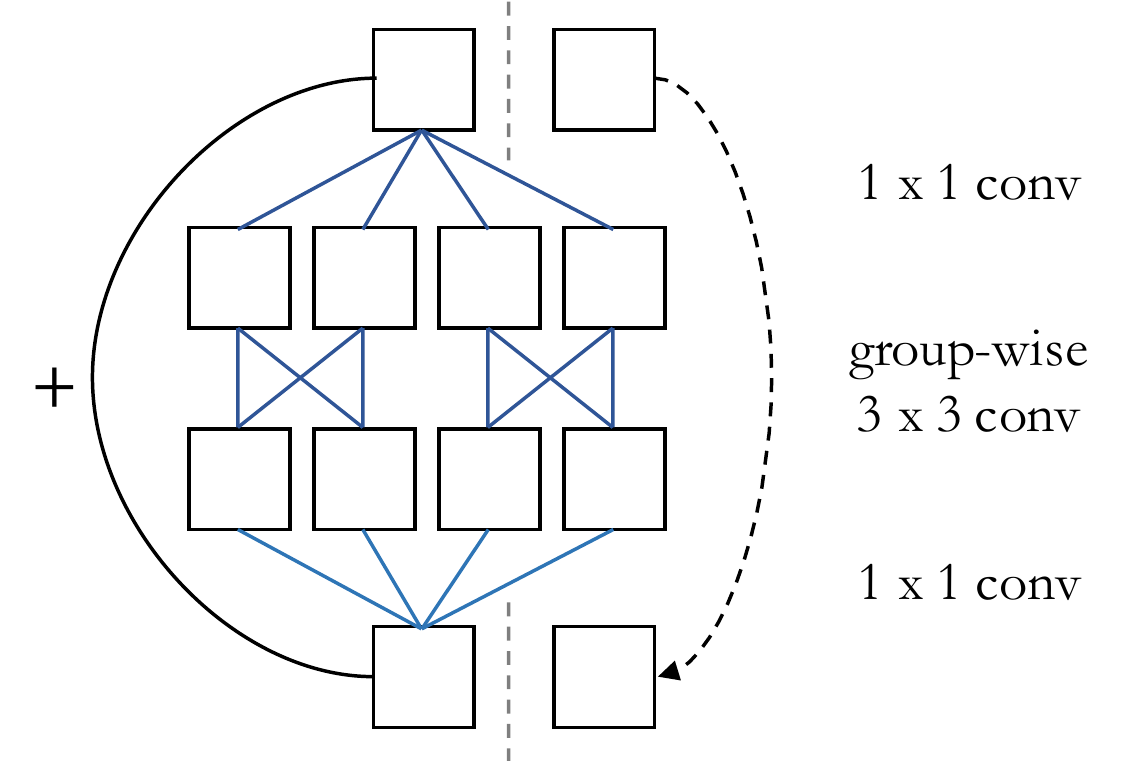}
		\caption{MUXConv block\label{fig:muxconv_block}\vspace{-0.3cm}}
	\end{subfigure}
	\caption{The visualization of the Mobilenet block (a) and our MUXConv block (b).
	\label{fig:block_comparison}\vspace{-0.3cm}}
\end{figure}

In this section, we analytically compare the computational complexity of our MUXConv block (Figure~\ref{fig:muxconv_block}) with the widely-used MobileNet block~\cite{mobilenetv2}. For simplicity, we ignore the computation induced by the normalization and activation layers and we assume that for both blocks the number of input and output channels is the same i.e., $C$ channels.

The Mobilenet block consist of a $1\times1$ convolution to expand the input channels, followed by a $3\times3$ depth-wise separable convolution and another $1\times1$ convolution to compress the channels (see Figure~\ref{fig:irb}). We use $E$ to denote expansion rate. Then the total number of parameters and floating point operations are:
\small{
\begin{align*}
\mbox{Params} & = \underbrace{C\cdot EC}_\textrm{$1\times1$ conv} + \underbrace{EC\cdot 3\cdot 3}_\textrm{$3\times3$ d.w. conv} + \underbrace{EC\cdot C}_\textrm{$1\times1$ conv} \\
\mbox{FLOPs} & =  H\cdot W\cdot \bigg(\underbrace{C\cdot EC\cdot}_\textrm{$1\times1$ conv} + \underbrace{EC\cdot 3\cdot 3}_\textrm{$3\times3$ d.w. conv} + \underbrace{EC\cdot C}_\textrm{$1\times1$ conv}\bigg)
\end{align*}}

On the other hand, our MUXConv block first select a subset of the input channels to be processed, and the remaining portion is directly propagated to the output. We use $L$ to denote the ratio of the leave-out un-processed channels. Then we use a $1\times1$ convolution to expand, followed by a group-wise convolution~\cite{resnext} and another $1\times1$ convolution to compress (see Figure~\ref{fig:muxconv_block}). And we use $G$ to denote the group factor, which indicates the \# of input channels used for computing each output channel. For instance, setting $G$ equal to 1 is equivalent as using a depth-wise separable convolution. The resulting number of parameters and the floating point operations associated with our MUXConv block is:
\small{
\begin{align*}
\hat{C} & = (1 - L)\cdot C \\
\mbox{Params} & = \underbrace{\hat{C}\cdot E\hat{C}}_\textrm{$1\times1$ conv} + \underbrace{G\cdot E\hat{C}\cdot 3\cdot 3}_\textrm{$3\times3$ group conv} + \underbrace{E\hat{C}\cdot \hat{C}}_\textrm{$1\times1$ conv} \\
\mbox{FLOPs} & =  H\cdot W\cdot \bigg(\underbrace{\hat{C}\cdot E\hat{C}\cdot}_\textrm{$1\times1$ conv} + \underbrace{G\cdot E\hat{C}\cdot 3\cdot 3}_\textrm{$3\times3$ group conv} + \underbrace{E\hat{C}\cdot \hat{C}}_\textrm{$1\times1$ conv}\bigg)
\end{align*}}

\begin{figure}[h]
	\centering
	\begin{subfigure}[t]{.45\textwidth}
		\centering
		\includegraphics[width=0.9\textwidth]{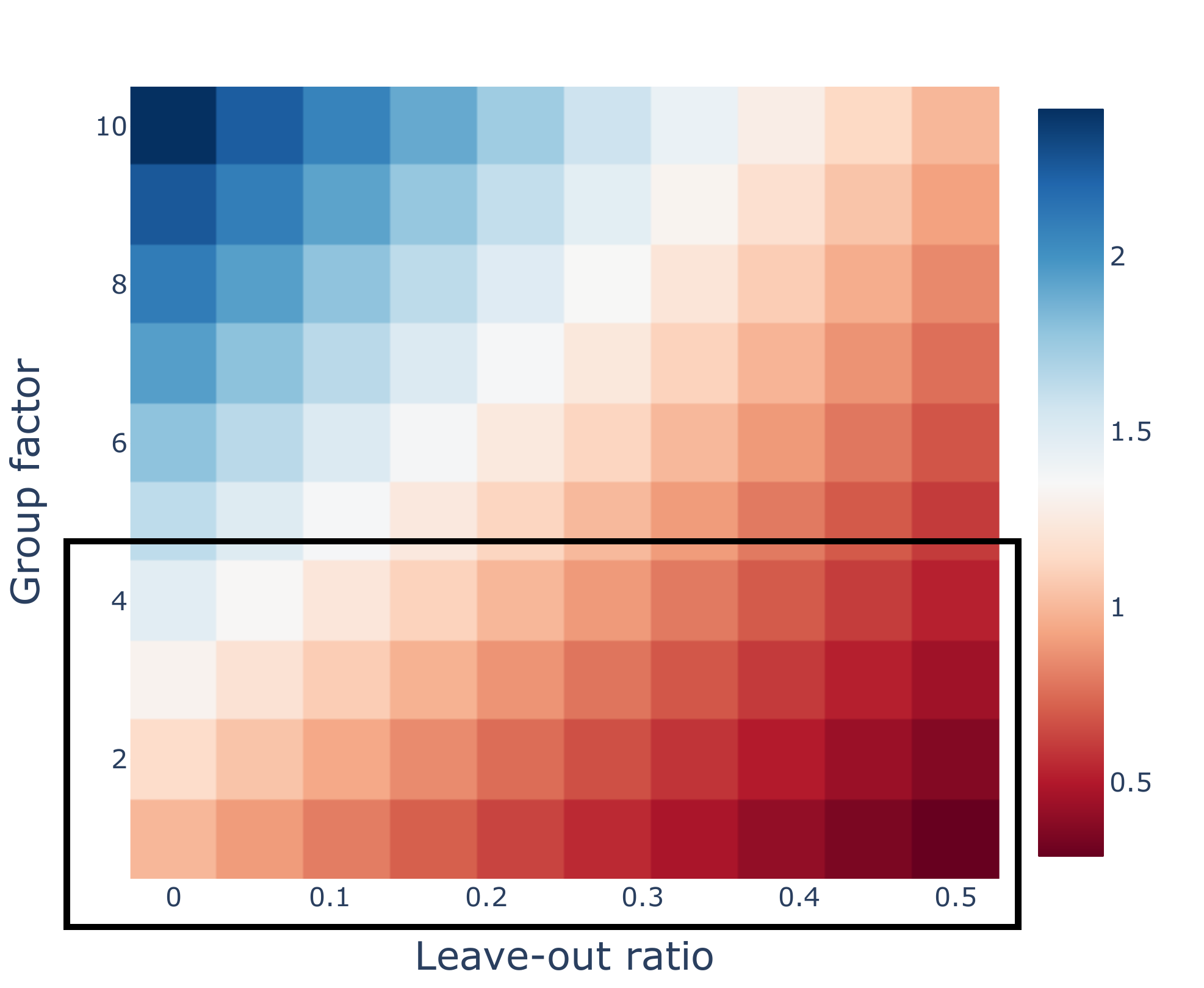}
	\end{subfigure}
	\caption{Ratio of \#Params between our MUXConv block and Mobilenet block~\cite{mobilenetv2}. The search space that we consider for these two hyperparameters is highlighted by a black box.
	\label{fig:computation_complexity}\vspace{-0.3cm}}
\end{figure}

Figure~\ref{fig:computation_complexity} provides an visual comparison showing the ratio of the number of parameters between our MUXConv block and Mobilenet block as the group factor ($G$) and leave-out ratio ($L$) vary. The choice of $G$ and $L$ hyperparameters we consider in our search space (see Table~\ref{tab:search-space}) corresponds to computational complexity that is less than the Mobilenet block (ratio $\leq$ 1, i.e. red color in Fig.\ref{fig:computation_complexity}).

\section{Additional Experiments}
\subsection{Semantic Segmentation\label{sec:semantic-segmentation}}
\begin{figure*}[ht]
	\centering
	\begin{subfigure}[b]{.95\textwidth}
		\centering
		\includegraphics[width=0.98\textwidth]{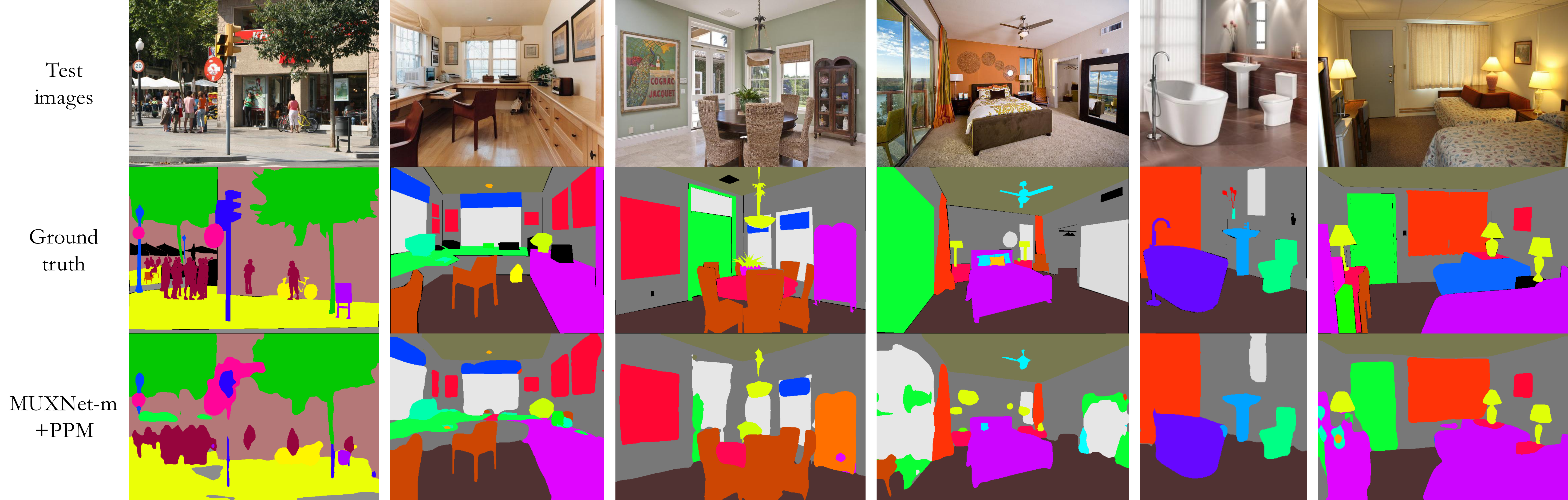}
	\end{subfigure}
	\caption{Examples from ADE20K validation set showing the ground truth (2nd row) and the scene parsing result (3rd row) from \ourmethod{}-m. Color encoding of semantic categories is available from \href{https://github.com/CSAILVision/semantic-segmentation-pytorch}{here}.
	\label{fig:segm_visualization}}
\end{figure*}

We further evaluate the effectiveness of our models as backbones for the task of mobile semantic segmentation. We compare \ourmethod{}-m with both MobileNetV2~\cite{mobilenetv2} and ResNet18~\cite{resnet} on ADE20K~\cite{zhou2019semantic} benchmark. Additionally, we also compare two different segmentation heads. The first one, referred as \emph{C1}, only uses one convolution module. And the other one, Pyramid Pooling Module (\emph{PPM}), was proposed in~\cite{zhao2017pspnet}. All models are trained under the same setup: we use SGD optimizer with initial learning rate 0.02, momentum 0.9, weight decay 1e-4 for 20 epochs. Table~\ref{tab:segmentation} reports the mean IoU (mIoU) and pixel accuracy on the ADE20K validation set. \ourmethod{}-m performs comparably with MobileNetV2 when paired with PPM, while being $1.5\times$ more efficient in MAdds. We also provide qualitative visualization of semantic segmentation examples in Figure~\ref{fig:segm_visualization}.

\begin{table}[t]
\centering
\resizebox{0.45\textwidth}{!}{%
\begin{tabular}{@{\hspace{2mm}}l|cc|cc@{\hspace{2mm}}}
\specialrule{1.5pt}{1pt}{1pt}
Network & \#MAdds & \#Params & mIoU (\%) & Acc (\%) \\
\specialrule{1.5pt}{1pt}{1pt}
ResNet18~\cite{resnet} + C1 & 1.8B & 11.7M & 33.82 & 76.05 \\
MobileNetV2~\cite{mobilenetv2} + C1 & 0.3B & 3.5M & 34.84 & 75.75 \\
\textbf{MUXNet-m + C1} & \textbf{0.2B} & \textbf{3.4M} & \textbf{32.42} & \textbf{75.00} \\ \midrule
ResNet18 + PPM & 1.8B & 11.7M & 38.00 & 78.64 \\
MobileNetV2 + PPM & 0.3B & 3.5M & 35.76 & 77.77 \\
\textbf{MUXNet-m + PPM} & \textbf{0.2B} & \textbf{3.4M} & \textbf{35.80} & \textbf{76.33} \\ \specialrule{1.5pt}{1pt}{1pt}
\end{tabular}%
}
\caption{ADE20K~\cite{zhou2019semantic} Semantic Segmentation Results. Since networks in each section use the same segmentation head, we report the \emph{\#MAdds} and \emph{\#Params} on the backbone models only. \emph{mIoU} is the mean IoU and \emph{Acc} is the pixel accuracy. \emph{C1} use one convolution module as segmentation head and \emph{PPM} use the Pyramid Pooling Module from~\cite{zhao2017pspnet}.
\label{tab:segmentation}\vspace{-0.3cm}}
\end{table}

\subsection{Generalization and Robustness \label{sec:generalization}}
\begin{figure*}[ht]
	\centering
	\begin{subfigure}[b]{.9\textwidth}
		\centering
		\includegraphics[width=0.98\textwidth]{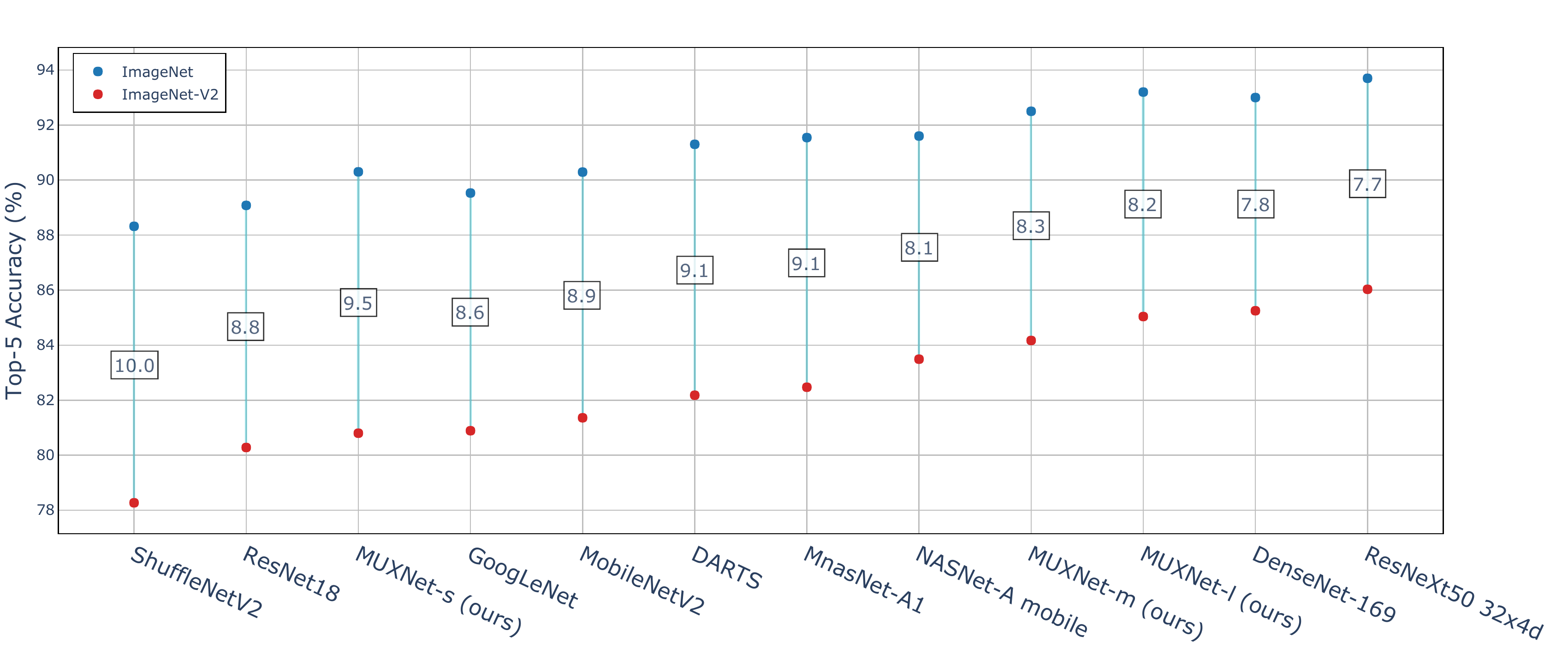}
		\caption{ImageNet-V2~\cite{imagenetv2} \label{fig:imagenetv2}\vspace{-0.1cm}}
	\end{subfigure}\\
	\begin{subfigure}[b]{.9\textwidth}
		\centering
		\includegraphics[width=0.98\textwidth]{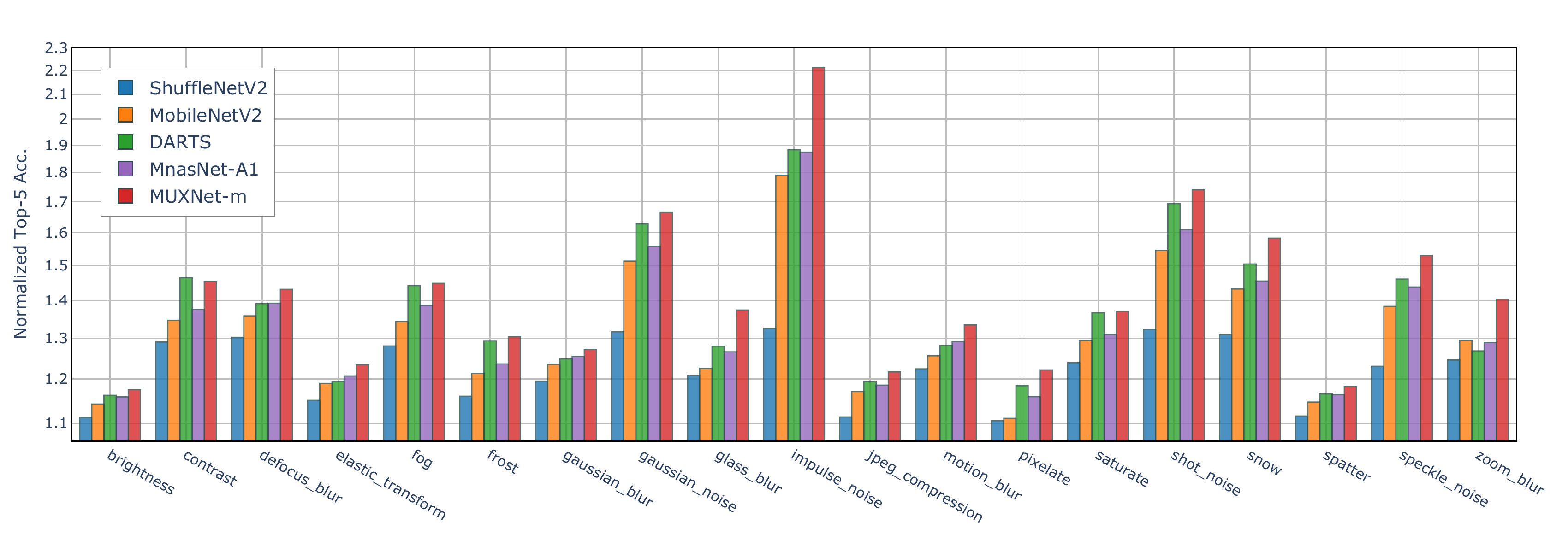}
		\caption{ImageNet-C~\cite{imagenet-c} \label{fig:imagenet-c}\vspace{-0.3cm}}
	\end{subfigure}\\
	\caption{(a) Generalization performance on ImageNet-V2 (MatchedFrequency)~\cite{imagenetv2}. Numbers in the boxes indicate the drop in accuracy. (b) Robustness performance on ImageNet-C~\cite{imagenet-c}, which consist of ImageNet validation images corrupted by 19 commonly observable corruptions. Following the original paper that proposed ImageNet-C, we normalized the top-5 accuracy by AlexNet's Top-5 accuracy. DARTS is from the author's public \href{https://github.com/quark0/darts}{Github repository}. All other compared models are from Pytorch repository \url{https://pytorch.org/docs/stable/torchvision/models.html}.\label{fig:general_robust}}
\end{figure*}

To further evaluate the generalization performance of our proposed models, we compare on a recently proposed benchmark dataset, ImageNetV2~\cite{imagenetv2}, complementary to the original ImageNet 2012. We use the \emph{MatchedFrequency} version of the ImageNet-V2. Figure~\ref{fig:imagenetv2} reports the top-5 accuracy comparison between our \ourmethod{}s and a wide-range of previous models. Even though there is a significant accuracy drop of 8\% to 10\% on ImageNet-V2 across models, the relative rank-order of accuracy on the original ImageNet validation set translates well to the new ImageNet-V2. And our \ourmethod{} performs competitively on ImageNet-V2 as compared to other mobile models, such as ShuffleNetV2~\cite{shufflenetv2}, MobileNetV2~\cite{mobilenetv2} and MnasNet-A1~\cite{mnasnet}.

The vulnerability to small changes in query images has always been a concern for designing better models. Hendrycks and Dietterich~\cite{imagenet-c} recently introduced a new dataset, ImageNet-C, by applying commonly observable corruptions (e.g., noise, weather, compression, etc.) to the clean images from the original ImageNet dataset. The new dataset contains images perturbed by 19 different types of corruption at five different levels of severity. And we leverage this dataset to evaluate the robustness of our proposed models. Figure~\ref{fig:imagenet-c} compares Top-5 accuracy between our \ourmethod{}-m and four other representative models, designed both manual and automatically. \ourmethod{}-m performs favourably on ImageNet-C, achieving better accuracy on 18 out of 19 corruption types.

\end{appendices}

{\small
\bibliographystyle{ieee_fullname}
\bibliography{egbib}
}

\end{document}